\documentclass{article}

\usepackage{microtype}
\usepackage{graphicx}
\usepackage{subcaption}
\usepackage[T1]{fontenc}
\usepackage{booktabs} 
\usepackage{longtable}
\usepackage{tcolorbox}
\newtcolorbox{mybody}{
  colback=myboxcolor,
  colframe=myframe,
  boxrule=1pt, 
  left=1pt,
  right=1pt,
  top=1pt,
  bottom=1pt,
}
\definecolor{myboxcolor}{RGB}{245,245,245} 
\definecolor{myframe}{RGB}{0,0,128} 

\definecolor{my_green}{rgb}{0.0, 0.6, 0.0}

\usepackage{hyperref}

\usepackage[accepted]{icml2025}

\usepackage{amsmath}
\usepackage{amssymb}
\usepackage{mathtools}
\usepackage{amsthm}
\DeclareMathOperator*{\argmax}{arg\,max}

\usepackage[capitalize,noabbrev]{cleveref}

\theoremstyle{plain}

\theoremstyle{definition}

\theoremstyle{remark}

\usepackage[textsize=tiny]{todonotes}

\icmltitlerunning{cMALC-D}

\begin{document}

\twocolumn[
\icmltitle{cMALC-D: Contextual Multi-Agent LLM-Guided Curriculum Learning with Diversity-Based Context Blending}




\icmlsetsymbol{equal}{*}

\begin{icmlauthorlist}
\icmlauthor{Anirudh Satheesh}{umd}
\icmlauthor{Keenan Powell}{umd}
\icmlauthor{Hua Wei}{asu}
\end{icmlauthorlist}

\icmlaffiliation{umd}{Department of Computer Science, University of Maryland, College Park, Maryland, USA}
\icmlaffiliation{asu}{School of Computing and Augmented Intelligence, Arizona State University, Tempe, Arizona, USA}

\icmlcorrespondingauthor{Anirudh Satheesh}{anirudhs@terpmail.umd.edu}
\icmlcorrespondingauthor{Keenan Powell}{kpowell1@terpmail.umd.edu}

\icmlkeywords{Machine Learning, ICML}

\vskip 0.3in
]



\printAffiliationsAndNotice{} 

\begin{abstract}
Many multi-agent reinforcement learning (MARL) algorithms are trained in fixed simulation environments, making them brittle when deployed in real-world scenarios with more complex and uncertain conditions. Contextual MARL (cMARL) addresses this by parameterizing environments with context variables and training a context-agnostic policy that performs well across all environment configurations. Existing cMARL methods attempt to use curriculum learning to help train and evaluate context-agnostic policies, but they often rely on unreliable proxy signals, such as value estimates or generalized advantage estimates that are noisy and unstable in multi-agent settings due to inter-agent dynamics and partial observability. To address these issues, we propose Contextual Multi-Agent LLM-Guided Curriculum Learning with Diversity-Based Context Blending (cMALC-D), a framework that uses Large Language Models (LLMs) to generate semantically meaningful curricula and provide a more robust evaluation signal. To prevent mode collapse and encourage exploration, we introduce a novel diversity-based context blending mechanism that creates new training scenarios by combining features from prior contexts. Experiments in traffic signal control domains demonstrate that cMALC-D significantly improves both generalization and sample efficiency compared to existing curriculum learning baselines. We provide code at \url{https://github.com/DaRL-LibSignal/cMALC-D}.
\end{abstract}

\section{Introduction}
Multi-Agent Reinforcement Learning (MARL) has shown promising results across diverse applications, including real-time strategy games \citep{samvelyan19smac, Kurach_Raichuk_Stańczyk_Zając_Bachem_Espeholt_Riquelme_Vincent_Michalski_Bousquet_Gelly_2020}, supply chain management \citep{liu2022multi, mousa2024analysis}, navigation and pathfinding \citep{zhang2024met, skrynnik2024learn}, and traffic signal control \citep{chu2019multi, jiang2022multi, satheesh2025constrained}. These successes are largely attributed to the ability of MARL algorithms, such as Independent Proximal Policy Optimization (IPPO) \citep{de2020independent} and Multi-Agent Proximal Policy Optimization (MAPPO) \citep{yu2022surprising}, to train agents capable of coordination and cooperation.

Despite this progress, generalization remains a key challenge. Most MARL algorithms are trained in simulation environments with fixed or limited variability, making them brittle when deployed in real-world scenarios where conditions are more complex and uncertain. External factors such as noise \citep{bukharin2023robust,he2023robust} and dynamic changes \citep{NEURIPS2020_77441296} can degrade MARL performance substantially. These issues are amplified in multi-agent settings due to the combinatorial explosion of agent interactions, which can destabilize learned policies and exacerbate overfitting to training conditions.
Prior work has addressed these challenges through meta-learning \citep{finn2017model, NEURIPS2023_d1b1a091, harris2022meta, zhang2021learning}, invariant representation learning \citep{chen2023rm, mcclellan2024boosting, mcclellan2025penguin}, and domain adaptation \citep{tzeng2020adapting, da2024prompt}. While effective, these approaches often rely on predefined environment distributions and cannot guarantee robustness to unseen or out-of-distribution scenarios.


To address the challenge of poor generalization to unseen or out-of-distribution environments, we build on the contextual MARL (cMARL) framework \citep{jayawardana2024intersectionzoo}, which explicitly represents environment variability through a context variable \(c\) \citep{hallak2015contextual}. Generalization in cMARL is commonly improved via curriculum learning \citep{bengio_cl}, where agents are trained on contexts that gradually increase in difficulty or novelty \citep{tobin2017domain, klink2020self, jiang2021prioritized, eimer2021self, parkerholder2023evolvingcurricularegretbasedenvironment}. This allows agents to acquire transferable skills incrementally and improves robustness at test time.

While curriculum learning improves generalization in contextual multi-agent reinforcement learning (MARL) by ordering training environment contexts by difficulty or novelty, existing approaches often depend on handcrafted heuristics or static curriculum schedules. These strategies may struggle to adapt to the evolving agents or the complex dependencies among context variables in dynamic environments.
To address these limitations, we explore the use of Large Language Models (LLMs) as high-level curriculum designers. Recent advancements in LLMs, such as GPT-4o, Qwen, and Gemini, have shown strong capabilities in reasoning, planning, and abstraction \citep{song2023llmplanner, yao2022webshop, xu2023llms, ma2023eureka, dong2024surveyincontextlearning}. These models can operate in diverse domains through in-context learning, suggesting their potential for adaptively generating and sequencing environment contexts based on the agent's current state and performance.

In this work, we propose Contextual Multi-Agent LLM-Guided Curriculum Learning with Diversity-Based Context Blending (cMALC-D), a novel framework that integrates LLMs into contextual MARL to dynamically generate training curricula. Specifically, in cMALC-D, the LLM acts as a high-level controller that observes the agent’s learning progress and adaptively proposes new environment contexts by reasoning over the space of context variables. To enhance coverage and prevent overfitting to narrow task distributions, we introduce a diversity-based context blending mechanism that mixes previously sampled contexts to construct novel yet meaningful training conditions. This LLM-guided process allows the curriculum to evolve in tandem with agent capabilities, providing more targeted and generalizable training experiences. We evaluate cMALC-D in multiple traffic control scenarios, where environments are naturally high-dimensional and dynamic. Results show that our approach significantly improves generalization to unseen environment contexts with higher sample efficiency compared to existing self-paced or handcrafted curriculum strategies.

Our contributions are outlined as follows:
\begin{itemize}
  \item We introduce Contextual Multi-Agent LLM-Guided Curriculum Learning with Diversity-Based Context Blending (cMALC-D), a framework that leverages Large Language Models (LLMs) to generate semantically meaningful context-based curricula for training MARL agents, improving generalization to unseen environment configurations.

  \item Experiments in multiple traffic-based environments demonstrate that our approach achieves better generalization compared to other self-paced curricula, with higher sample efficiency.
\end{itemize}


\section{Related Work}
\subsection{Generalization in MARL}
\textbf{Meta Learning} One approach to improving MARL generalizability has been to apply meta-learning to MARL. Meta learning develops a good policy initialization that can rapidly adapt to new tasks or environmental configurations with minimal additional training. This can be done with Model-Agnostic Meta-Learning (MAML) algorithms \citep{finn2017model, NEURIPS2023_d1b1a091} or through context-based latent variables \citep{wen2024dream, zhang2021learning}, both of which train over a task distribution to develop transferable inductive biases. However, such methods face significant scalability issues. MAML requires expensive Hessian computations on the order of the number of model parameters or first-order approximations \citep{fallah2021convergence, nichol2018first}. Additionally, prior work focuses on multi-agent solution concepts for two-player games in tabular settings \citep{NEURIPS2023_d1b1a091, harris2022meta, pmlr-v162-zhang22an}, which are not extendable to many agents parameterized by neural networks.

\textbf{Equivariance} Multi-agent environments exhibit symmetries that can be leveraged to improve generalization using equivariant networks \citep{mcclellan2024boosting, mcclellan2025penguin, chen2023rm}. To fully exploit such invariances, each agent must be parameterized with architectures that respect these symmetries, such as EGNNs \citep{pmlr-v139-satorras21a}, E3-MPNNs \citep{brandstetter2021geometric}, or E2GN2s \citep{mcclellan2024boosting}. However, these methods typically require access to the full global state, limiting their applicability in partially observable settings, which encompass most multi-agent environments.

\textbf{Sim-to-Real}
It is often very challenging to train MARL agents on real-world environment data due to limited accessibility, high costs, and the intricate complexities of real-world systems. A common workaround is to train agents in simulated environments instead. However, simulations often fail to capture the full range of noise and unpredictability found in the real world, making it difficult for agents to transfer their skills effectively \citep{Dulac-Arnold2021}. Several techniques have been proposed to address this sim-to-real gap, one notable method being Domain Randomization. Domain Randomization exposes agents to a wide range of settings by randomizing key environment parameters \citep{9308468}. While effective in some cases, it can produce unrealistic environments that fail to teach agents transferable skills, an issue that becomes more pronounced in MARL settings. Additional methods try to minimize the sim-to-real gap by aligning simulation trajectories to real ones \citep{tzeng2020adapting}, or by developing an inverse model to correct for bias in the simulation environment \citep{da2024prompt}, but these methods require information from specific real-world environments, which may not be available. Recently, large language model (LLM)-based methods have shown promise in generating more coherent and purposeful environments, but their application has been largely limited to single-agent contexts \citep{ma2024dreurekalanguagemodelguided, zala2024envgengeneratingadaptingenvironments}.

\subsection{Self-Paced Curriculum Learning}
In contrast to prior approaches such as meta-learning, equivariant architectures, and sim-to-real, curriculum learning \citep{bengio_cl} aims to improve generalization by structuring agents' learning process, progressing from easier to more challenging tasks. Additional methods explore automated curriculum generation to avoid manual design. For example, \citet{sukhbaatar2017} leverages self-play to minimize the number of training episodes by generating progressively harder tasks through agent interactions. \citet{dendorfer2020goal} and \citet{florensa2018automatic} use Generative Adversarial Networks (GANs) to create challenging goals tailored to the agent's capabilities. \citet{portelas2020teacher} uses a Gaussian Mixture Model (GMM) to model the task space and align a student’s learning trajectory with a teacher-generated curriculum. However, each of these requires an auxiliary model to determine the learnability of a task. Instead, \citet{klink2020self, eimer2021self} use self-paced learning, which uses agents' performance to order the curriculum. Thus, each task is tailored to each agent's abilities and ensures that the learning progress is more self-contained. 

\section{Contextual Multi-Agent Reinforcement Learning}
We formulate cMALC-D as a \textit{contextual decentralized partially observable Markov decision process} (cDec-POMDP). A cDec-POMDP is parameterized by the tuple \(M_c = (\mathcal{N}, \mathcal{S}, \mathcal{A}, \mathcal{T}_c, R, \Omega, \mathcal{O}, \gamma, \mu)\), where \(\mathcal{N}\) is the set of agent indices, denoting a system of \(n = |\mathcal{N}|\) cooperative agents. \(\mathcal{S}\) is the joint state space shared across all agents, and \(\mathcal{A} = \prod_{i \in \mathcal{N}} \mathcal{A}^i\) is the joint action space, where \(\mathcal{A}^i\) is the action space for agent \(i\). The transition function \(\mathcal{T}_c: \mathcal{S} \times \mathcal{A} \rightarrow \Delta(\mathcal{S})\) determines the next state distribution given the current state and joint action under context \(c\), while the reward function \(R: \mathcal{S} \times \mathcal{A} \rightarrow \mathbb{R}\) maps state-action pairs to a scalar reward. \(\Omega = \prod_{i \in \mathcal{N}} \Omega^i\) denotes the joint observation function, where \(\Omega^i: \mathcal{S} \rightarrow \mathcal{O}^i\) provides a private observation to agent \(i\), and \(\mathcal{O} = \prod_{i \in \mathcal{N}} \mathcal{O}^i\) is the joint observation space. The discount factor \(\gamma \in [0,1)\) specifies the importance of future rewards, and \(\mu: \mathcal{S} \rightarrow [0,1]\) is the initial state distribution.

To model task variation, we define a distribution over contexts \(c \in \mathcal{C}\), where each context specifies a different task instance by altering the transition functions. This induces a set \(\mathcal{M}_{\mathcal{C}} = \{M_c | c \in \mathcal{C}\}\) of decentralized POMDPs, each corresponding to a distinct environment. Each \(M_c\) encodes a different task instantiation with a different transition function, and we assume that the reward function and the state, action, and observation spaces remain fixed across all contexts.

The objective of a policy \(\pi\) in a cDec-POMDP is to maximize the expected return over the context distribution and over finite horizon \(H\):
\begin{align*}
\pi^* = \argmax_{\pi \in \Pi} \mathbb{E}_{c \sim \mathcal{C}} \left[ \mathbb{E}_{\pi} \left[ \sum_{t=0}^{H-1} \gamma^t \mathcal{R}(s_t, a_t) {\bigg|} \pi, c \right] \right]
\end{align*}
where \(\Pi = \{\pi = (\pi^1, \ldots, \pi^n) \mid \pi^i : o^i_t \rightarrow a^i_t\}\) is the set of decentralized policies, with each agent \(i \in \mathcal{N}\) selecting actions based only on its local observation \(o^i_t\). The goal is to learn a context-agnostic policy that generalizes well across the set of Dec-POMDPs \(\mathcal{M}_{\mathcal{C}}\).

\section{Evolutionary LLM Self-Paced Curriculum}
In this section, we outline the motivations of CMALC-D and its implementation.

\subsection{Limitations of Existing Self-Paced Curriculum Learning Algorithms}
There are two major limitations of current self-paced curriculum learning algorithms for contextual MARL (cMARL), both of which hinder efficient and robust generalization:
\begin{itemize}
  \item \textbf{Random Task Sampling} Most existing approaches generate new contexts by randomly sampling from the context space, without considering semantic relationships or difficulty progression between sampled environments. This can lead to large variations between contexts across training episodes, making learning unstable and inefficient. For example, an agent might face an easy environment followed by a drastically more difficult or qualitatively different one, forcing it to relearn strategies rather than incrementally building off prior knowledge.
  \item \textbf{Unreliable Proxy Evaluation} Current methods typically rely on policy metrics like the value estimate or the Generalized Advantage Estimate (GAE) to evaluate agent performance and determine subsequent contexts to train on. While these methods work well for single-agent tasks with many environment updates \citep{jiang2021prioritized, parkerholder2023evolvingcurricularegretbasedenvironment}, these metrics can be unreliable during early training phases and under sudden domain transfer. They can also be overconfident in underexplored regions of the context space and may not provide a meaningful indication of how well agents will generalize to unseen contexts. 
\end{itemize}

These challenges underscore the need for more structured, semantically aware curriculum generation strategies coupled with more robust evaluation signals that better reflect generalization and learning progress across context variations.

\subsection{Evolutionary LLM-Guided Curriculum for Context Generation}

To address these limitations, we propose cMALC-D, a novel curriculum learning strategy for contextual MARL that combines the structured reasoning capabilities of large language models (LLMs) with an exploration mechanism based on task arithmetic. This approach improves both the generation of semantically meaningful environment contexts and the robustness of policy evaluation under limited feedback.

\textbf{LLM-Guided Context Generation} Instead of randomly sampling from a context space \(C\), we leverage a large language model to reason over a sliding window of past training results and generate new contexts that reflect a meaningful progression in difficulty or diversity. At each curriculum step, the LLM receives a window of the most recent contexts \(\{c_{t-w}, \cdots, c_{t}\}\) and their associated performance metrics \(\{m_{t-w}, \cdots m_{t}\}\) when trained on MARL algorithm \(A\). It then leverages this history to propose a new context that either incrementally challenges the current multi-agent policy or targets known weaknesses observed in recent episodes.

\textbf{Diversity-Based Context Blending}
To avoid curriculum stagnation and encourage exploration of the context space, we monitor the similarity between successive contexts. If the LLM repeatedly generates highly similar contexts, indicating potential mode collapse in curriculum progression, we enable a diversity mechanism. Specifically, when the number of consecutive similar tasks exceeds a threshold, we blend the current LLM-proposed context with a randomly sampled context from history. This interpolation helps inject novelty in the curriculum while avoiding sudden changes in the curriculum.

\textbf{Alternating Policy Training and Context Generation} Similar to \citep{ma2023eureka}, we alternate between policy training and context generation. After each training phase, the agent's performance on the current context is recorded and passed to the LLM, which conditions on a sliding window of past evaluations to generate the next context. This approach—in-context context generation—enables the LLM to implicitly reason about task difficulty and progression without gradient updates or handcrafted reward shaping.

We present the full algorithm in Algorithm \(\ref{alg:llm_diverse_curriculum}\). 

\begin{algorithm}[h]
\caption{Contextual Multi-Agent LLM-Guided Curriculum Learning with Diversity Based Context Blending (cMALC-D)}
\label{alg:llm_diverse_curriculum}
\begin{algorithmic}[1]
\REQUIRE MARL algorithm \(A\), context space \(C\), LLM \(M\), blending factor \(\alpha\), sliding window size \(w\), similarity threshold \(\delta\), max similar count \(k\), initial context \(c_0\)
\STATE Initialize context buffer \(H \leftarrow []\), similarity counter \(s \leftarrow 0\)
\STATE Set current context \(c_0\)
\FOR{curriculum step \(t = 0, 1, \dots, T\)}
  \STATE Train policy \(\pi_t\) on context \(c_t\), collect performance metric \(m_t\)
  \STATE Append \((c_t, m_t)\) by algorithm \(A\) to context buffer \(H\)
  \STATE Construct sliding window \(H_w = \{(c_{t-w}, m_{t-w}), \dots, (c_t, m_t)\}\)
  \STATE Query \(M\) with \(H_w\) to generate new context \(c_{t+1}^{M}\)
  \STATE Compute similarity \(\sigma \leftarrow \text{Sim}(\{c_{t-w}, \cdots, c_t\}, c_{t+1}^{M})\)
  \IF{\(\sigma \geq \delta\)}
    \STATE Increment similarity counter \(s \leftarrow s + 1\)
  \ELSE
    \STATE Reset similarity counter \(s \leftarrow 0\)
  \ENDIF
  \IF{\(s \geq k\)}
    \STATE Sample random prior context \(c_r \sim \text{Uniform}(H)\)
    \STATE Blend: \(c_{t+1} \leftarrow \alpha c_r + (1 - \alpha) c_{t+1}^{M}\)
    \STATE Reset similarity counter \(s \leftarrow 0\)
  \ELSE
    \STATE Set \(c_{t+1} \leftarrow c_{t+1}^{M}\)
  \ENDIF
\ENDFOR
\end{algorithmic}
\end{algorithm}

\section{Experimental Setup}
\label{sec: experimental setup}
In this section, we show experimental details and the baselines we evaluate cMALC-D against.

\subsection{Experimental Details}
\label{subsec: experimental details}
We evaluate cMALC-D on three autonomous traffic signal control environments based on real-world data. We choose to evaluate on these datasets for three reasons:
\begin{itemize}
  \item \textbf{Real-World Relevance} Autonomous traffic signal control has critical real-life applications in reducing congestion, emissions, and travel time. This also gives several metrics outside of reward that we can use for evaluation. More information about environment metrics can be found in Appendix \ref{subsec: environment details}
  \item \textbf{Well Defined Context Space} Each vehicle in the traffic signal control environment is clearly defined by several variables that can be altered to generate new contexts. More details about context parameterization can be found in Appendix \ref{subsec: context parameterization}.
  \item \textbf{Multi-Agent Coordination Challenges} Traffic signal control inherently involves decentralized decision-making and coordination among agents, providing a natural testbed for evaluating the scalability and generalization of our cMARL curriculum framework.
\end{itemize}

We run our experiments with the CityFlow environment, which is a realistic and efficient traffic flow simulator written in C++ \citep{zhang2019cityflow}. It is also compatible with MARL algorithms via integration with the Gymnasium Library \citep{brockman2016openai}. We train all policies with MAPPO \citep{yu2022surprising}, but any MARL algorithm will work; we choose to use MAPPO due to its efficiency compared to off-policy algorithms.

We use the Qwen2.5-7B-Instruct model \citep{qwen2.5} as a high-level curriculum designer. To increase computational efficiency, we use the vLLM package \citep{kwon2023efficient} to streamline inference and leverage activation-aware weight quantization \citep{lin2024awq} to reduce memory usage. Thus, we only require 2 RTX 2080 TI GPUs for LLM-based experiments, and 1 RTX 2080 TI GPU for non-LLM experiments, and there is a neglible difference in MARL policy training time.

For all experiments, we alternate between expanding the curriculum and training the MARL policy for 500 episodes, where each episode is 360 timesteps, resulting in 180,000 trajectories per training phase. This is done via the Centralized Training Decentralized Execution (CTDE) paradigm, where training is done with access to the global joint observation, but agents only have access to their local observations during policy execution. We reserve a held-out test set of 5 contexts and evaluate the current policy every 5 episodes using greedy action sampling. After training, we generate 10 additional random contexts to assess generalization performance in both a zero-shot setting and after a brief finetuning phase of 5 episodes. In practice, we observe minimal differences between the zero-shot and finetuned policies; therefore, we report performance metrics, such as delay time, throughput, and wait time, based on the finetuned policy. Additional details about these metrics can be found in Appendix \ref{sec: additional experiments details}. All experiments are repeated across 5 random seeds to ensure statistical robustness.

\subsection{Baselines}
In this work, we compare cMALC-D to 5 other curriculum learning algorithms. 
\begin{itemize}
  \item \textbf{No Curriculum}: For this algorithm, we simply train MARL agents using MAPPO on a single training environment for all epochs, as would be done normally.
  \item \textbf{Domain Randomization} \citep{tzeng2020adapting}: Domain Randomization randomly generates an environment using a set of parameters and a probability distribution for each parameter at each timestep. Then, a policy is trained on a rollout from that environment at each timestep. 
  \item \textbf{Prioritized Level Replay (PLR)} \citep{jiang2021prioritized}: PLR is a curriculum learning algorithm that keeps track of previously generated environments (or levels) and scores them based on learning potential using the Temporal Difference Error. After evaluating each level, it decides randomly whether to replay the previously played level with the highest learning potential or create a new level. 
  \item \textbf{Adversarially Compounding Complexity by Editing Levels (ACCEL)} \citep{parkerholder2023evolvingcurricularegretbasedenvironment}: ACCEL is similar to PLR, except instead of training directly on previous levels with high TD-error, ACCEL maintains a population of levels with high TD-error and makes small and randomized mutations to those levels, and decides randomly whether to play on one of the population of those levels or generate a new one.
  \item \textbf{Self-Paced Context Evaluation (SPACE)} \citep{eimer2021self}: SPACE uses the value estimate to build a curriculum. Taking $V^\pi_t(s_0, c_i)$ to be the estimated total expected reward by following policy $\pi$ from starting state $s_0$ with context $c_i$ after $t$ training steps, SPACE takes $V^\pi_t(s_0, c_i)- V^\pi_{t-1}(s_0, c_i)$ to be the performance improvement capacity (PIC) for \(c_i\). SPACE trains continually in environments with high PIC until it converges, sampling a new environment from the context space.
\end{itemize}

\section{Results}
In this section, we describe the results of cMALC-D compared to the baselines and perform ablation studies to understand individual components of the algorithm. We aim to answer three main questions: What is the generalization performance of the algorithm? How does the diversity mechanism influence context generation? What kinds of contexts are generated by the LLM curriculum?

\subsection{Generalization Performance}
We show the generalization performance of cMALC-D against the baseline algorithms in Tables \ref{tab:after_finetune_cityflow1x3}, \ref{tab:after_finetune_HZ}, and \ref{tab:after_finetune_JN}. Across all three environments, JN \(1 \times 3\), HZ, and JN \(3 \times 4\), cMALC-D consistently outperforms or matches all other curriculum strategies on the test reward and specific traffic policy metrics, such as average delay and throughput. For example, in JN \(1 \times 3\), it achieves the highest test reward (\(29.01 \pm 0.32\)) and throughput (\(3073.22 \pm 114.06\)) while reducing wait time by \(2\%\) over the second-best algorithm. Similar trends hold for the HZ and JN \(3 \times 4\) environments.

\textbf{Structured curricula are necessary to learn generalizable policies.} In contrast, Domain Randomization underperforms compared to cMALC-D, often giving 3rd or 4th place results across performance metrics (e.g., 4th place in average delay in HZ with $241.79 \pm 35.60$ vs. cMALC-D's $146.96 \pm 19.93$). While it occasionally yields high throughput or test rewards over other algorithms (e.g., 2nd place test reward of $27.55 \pm 0.41$ in JN $1 \times 3$), these gains are unreliable and highly environment-dependent. This inconsistency highlights a fundamental limitation of randomization-based strategies: while they expose agents to a wide range of environments, they do so without considering progression or context relevance. As a result, agents may struggle to learn the high-level coordination skills necessary for generalization due to rapid context switching in the curriculum.

\textbf{Original context can be a useful prior, but may encourage overfitting.} Training without a curriculum can yield strong performance, particularly in the JN environments, where No Curriculum frequently ranks second after cMALC-D (e.g., throughput of $3704.17 \pm 147.39$ vs. cMALC-D's $3795.46 \pm 159.50$ in JN $3 \times 4$). This suggests that the original context provides a good prior, enabling agents to learn basic coordination strategies. However, its effectiveness diminishes in more diverse settings (most notably in the HZ environment, where its average delay of $339.09 \pm 53.28$ is worse than cMALC-D's $146.96 \pm 19.93$), where it performs significantly worse than cMALC-D and exhibits high variance even in the JN environments (e.g., test reward standard deviation of $2.44$ vs. $1.53$ in JN $3 \times 4$). This drop indicates that without curriculum learning, agents may overfit to features in the original context, which limits generalizability.

\textbf{LLM-based context evaluation provides a stable signal for effective curriculum learning.} While some methods like ACCEL and PLR incorporate similar automatic curriculum schemes, they rely heavily on policy evaluation signals, such as value functions or generalized advantage estimates, to select and schedule tasks. While these signals can be highly effective in single-agent domains with millions of environment updates, they can be noisy or unreliable in MARL due to non-stationarity, partial observability, and inter-agent dependencies (e.g., ACCEL's inconsistent rankings from 4th place in JN $1 \times 3$ to 2nd place in HZ). On the other hand, cMALC-D's context selection strategy promotes gradual skill acquisition that transfers well across diverse contexts. This is due to using language-based evaluations that can capture qualitative improvements that traditional metrics might overlook (demonstrated by cMALC-D's top performance across all environments with test rewards of $29.01 \pm 0.32$, $172.87 \pm 1.03$, and $116.57 \pm 1.53$ in JN $1 \times 3$, HZ, and JN $3 \times 4$, respectively).

\begin{table*}[ht!]
\centering
\caption{Performance metrics across all environments. Best results per metric are shown in \textbf{bold} and second-best results are \underline{underlined}. We include uncertainty within one standard deviation of the mean, averaged over 5 seeds.}
\label{tab:all_results}

\begin{subtable}{\textwidth}
\begin{tabular}{lccccc}
\toprule
Curriculum & Average Time & Throughput & Average Wait Time & Average Delay & Test Reward \\
\midrule
No Curriculum & \underline{816.00} $\pm$ \underline{35.63} & \underline{3032.72} $\pm$ \underline{114.20} & \underline{765.84} $\pm$ \underline{36.76} & \underline{728.34} $\pm$ \underline{37.56} & $27.56 \pm 0.46$ \\
Domain Randomization & $865.64 \pm 43.08$ & $2807.22 \pm 141.63$ & $816.64 \pm 44.41$ & $774.43 \pm 46.66$ & 27.55 $\pm$ 0.41 \\
PLR & $841.74 \pm 40.63$ & $2955.79 \pm 127.53$ & $795.98 \pm 41.65$ & $772.72 \pm 41.46$ & $\underline{27.61} \pm \underline{0.44}$ \\
ACCEL & $860.81 \pm 39.57$ & $2813.86 \pm 130.81$ & $819.49 \pm 40.75$ & $806.07 \pm 41.06$ & $26.55 \pm 0.45$ \\
SPACE & $939.39 \pm 39.96$ & $2544.24 \pm 133.01$ & $898.71 \pm 41.15$ & $875.71 \pm 41.19$ & $26.89 \pm 0.43$ \\
cMALC-D & \textbf{809.39} $\pm$ \textbf{36.37} & \textbf{3073.22} $\pm$ \textbf{114.06} & \textbf{750.81} $\pm$ \textbf{36.99} & \textbf{718.77} $\pm$ \textbf{37.26} & \textbf{29.01} $\pm$ \textbf{0.32} \\
\bottomrule
\end{tabular}
\caption{JN \(1 \times 3\) Performance Metrics}
\label{tab:after_finetune_cityflow1x3}
\end{subtable}

\begin{subtable}{\textwidth}
\begin{tabular}{lccccc}
\toprule
Curriculum & Average Time & Throughput & Average Wait Time & Average Delay & Test Reward \\
\midrule
No Curriculum & $710.01 \pm 42.78$ & $2188.94 \pm 74.78$ & $342.84 \pm 51.55$ & $339.09 \pm 53.28$ & $164.04 \pm 2.68$ \\
Domain Randomization & $637.52 \pm 31.61$ & $2318.41 \pm 53.49$ & $239.36 \pm 32.60$ & $241.79 \pm 35.60$ & $168.53 \pm 1.44$ \\
PLR & $611.26 \pm 26.48$ & $2390.56 \pm 38.11$ & $199.22 \pm 23.74$ & $175.96 \pm 22.20$ & $171.85 \pm 1.02$ \\
ACCEL & 615.02 $\pm$ 25.46 & $2393.30 \pm 35.37$ & $212.28 \pm 21.12$ & $178.54 \pm 19.56$ & $171.07 \pm 1.17$ \\
SPACE & $\underline{588.84} \pm \underline{25.11}$ & \textbf{2440.18} $\pm$ \textbf{32.02} & \underline{166.75} $\pm$ \underline{18.13} & \textbf{140.25} $\pm$ \textbf{16.30} & \textbf{172.90} $\pm$ \textbf{1.05} \\
cMALC-D & \textbf{586.86} $\pm$ \textbf{25.50} & \underline{2440.09} $\pm$ \underline{35.45} & \textbf{163.92} $\pm$ \textbf{19.71} & \underline{146.96} $\pm$ \underline{19.93} & \underline{172.87} $\pm$ \underline{1.03} \\
\bottomrule
\end{tabular}
\caption{HZ Performance Metrics}
\label{tab:after_finetune_HZ}
\end{subtable}

\begin{subtable}{\textwidth}
\begin{tabular}{lccccc}
\toprule
Curriculum & Average Time & Throughput & Average Wait Time & Average Delay & Test Reward \\
\midrule
No Curriculum & \underline{829.92} $\pm$ \underline{36.98} & \underline{3704.17} $\pm$ \underline{147.39} & \underline{566.12} $\pm$ \underline{43.18} & \underline{550.47} $\pm$ \underline{46.03} & \underline{115.77} $\pm$ \underline{2.44} \\
Domain Randomization & $976.73 \pm 37.67$ & $3108.67 \pm 151.36$ & $753.96 \pm 45.58$ & $756.78 \pm 49.21$ & $112.38 \pm 1.52$ \\
PLR & $992.79 \pm 47.30$ & $3071.38 \pm 188.12$ & $781.35 \pm 57.70$ & $786.23 \pm 61.33$ & $111.98 \pm 1.74$ \\
ACCEL & $1077.27 \pm 42.13$ & $2699.78 \pm 166.09$ & $903.94 \pm 50.26$ & $921.75 \pm 52.73$ & $110.28 \pm 1.60$ \\
SPACE & $899.94 \pm 43.34$ & $3447.63 \pm 172.27$ & $658.07 \pm 51.30$ & $638.21 \pm 55.47$ & $114.47 \pm 1.64$ \\
cMALC-D & \textbf{815.81} $\pm$ \textbf{39.51} & \textbf{3795.46} $\pm$ \textbf{159.50} & \textbf{557.96} $\pm$ \textbf{46.57} & \textbf{529.83} $\pm$ \textbf{50.28} & \textbf{116.57} $\pm$ \textbf{1.53} \\
\bottomrule
\end{tabular}
\caption{JN \(3 \times 4\) Performance Metrics}
\label{tab:after_finetune_JN}
\end{subtable}
\end{table*}

\subsection{Influence of the Diversity Mechanism}
To evaluate the impact of the diversity mechanism, we compare three variants of our method: the full version with similarity-based diversity (\textbf{cMALC-D}), a baseline without the diversity mechanism (\textbf{cMALC}), and a variant that applies task arithmetic with random probability \(\epsilon = 0.1\) instead of using similarity checks (\textbf{cMALC-\(\epsilon\)}).

\begin{figure*}[ht!]
  \centering
  \begin{subfigure}[b]{0.32\textwidth}
    \includegraphics[width=\linewidth]{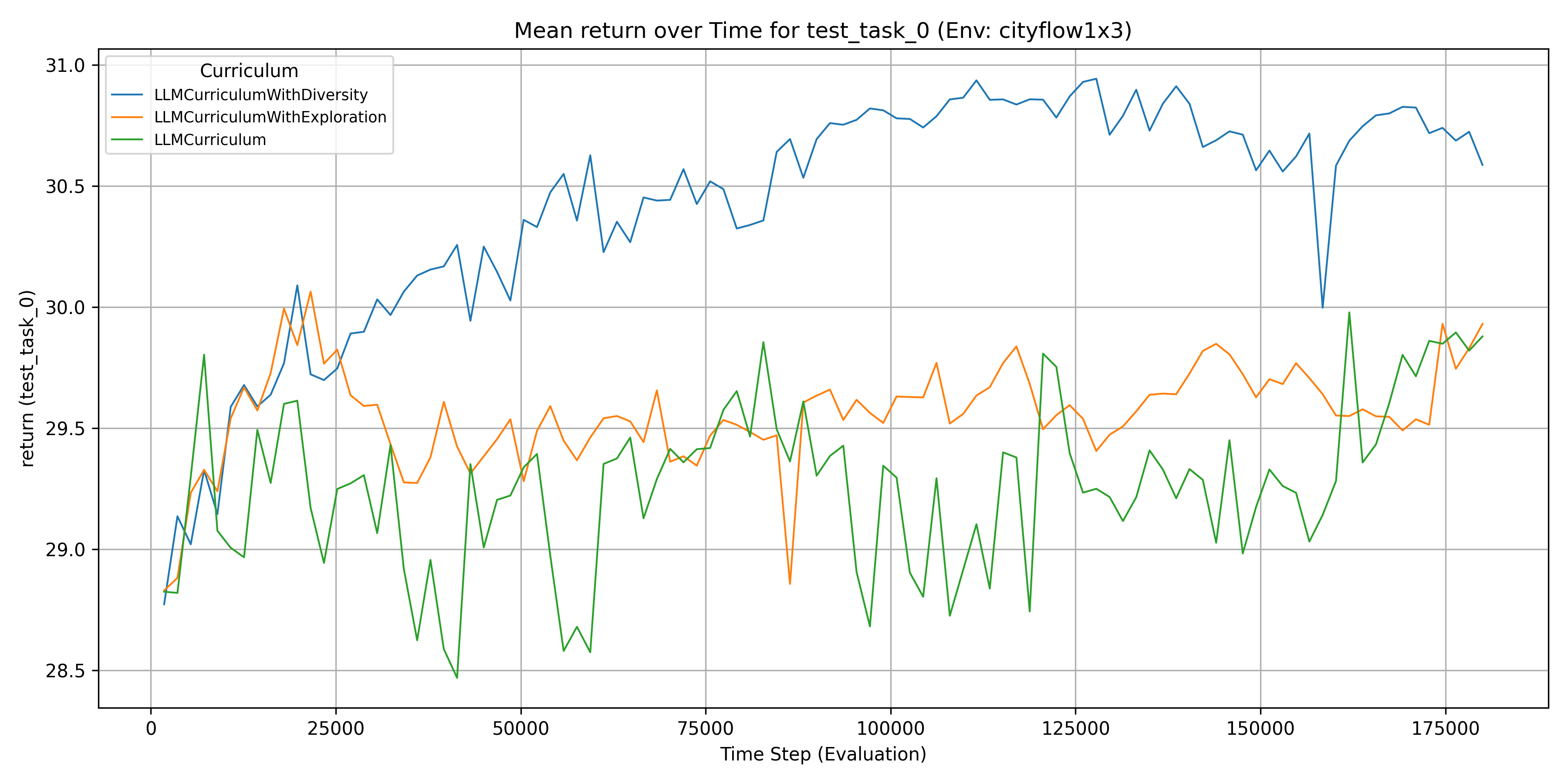}
    \caption{Jinan (1×3), Task 1}
    \label{fig:task0_cityflow1x3}
  \end{subfigure}%
  \begin{subfigure}[b]{0.32\textwidth}
    \includegraphics[width=\linewidth]{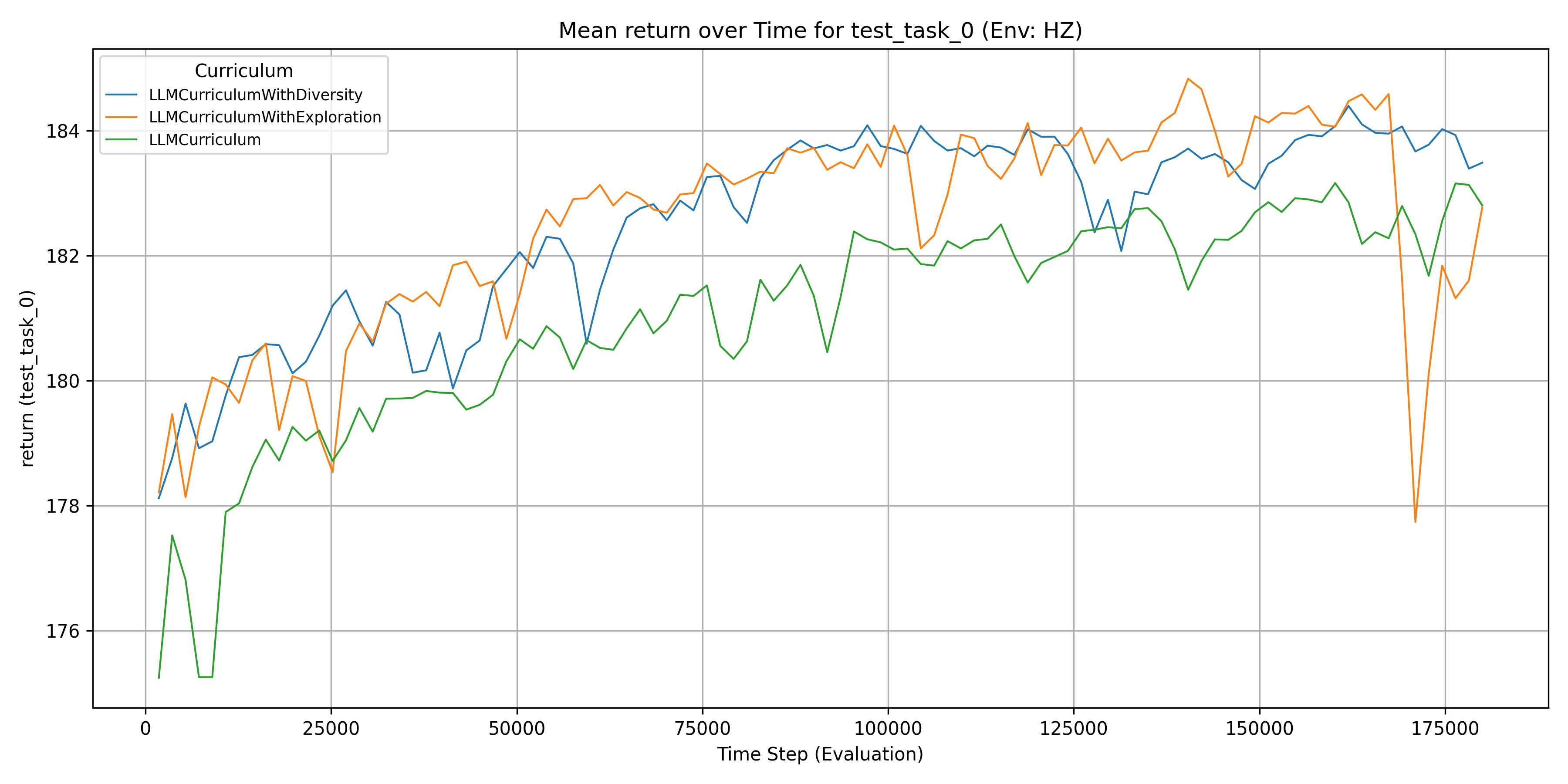}
    \caption{Hangzhou, Task 1}
    \label{fig:task0_HZ}
  \end{subfigure}%
  \begin{subfigure}[b]{0.32\textwidth}
    \includegraphics[width=\linewidth]{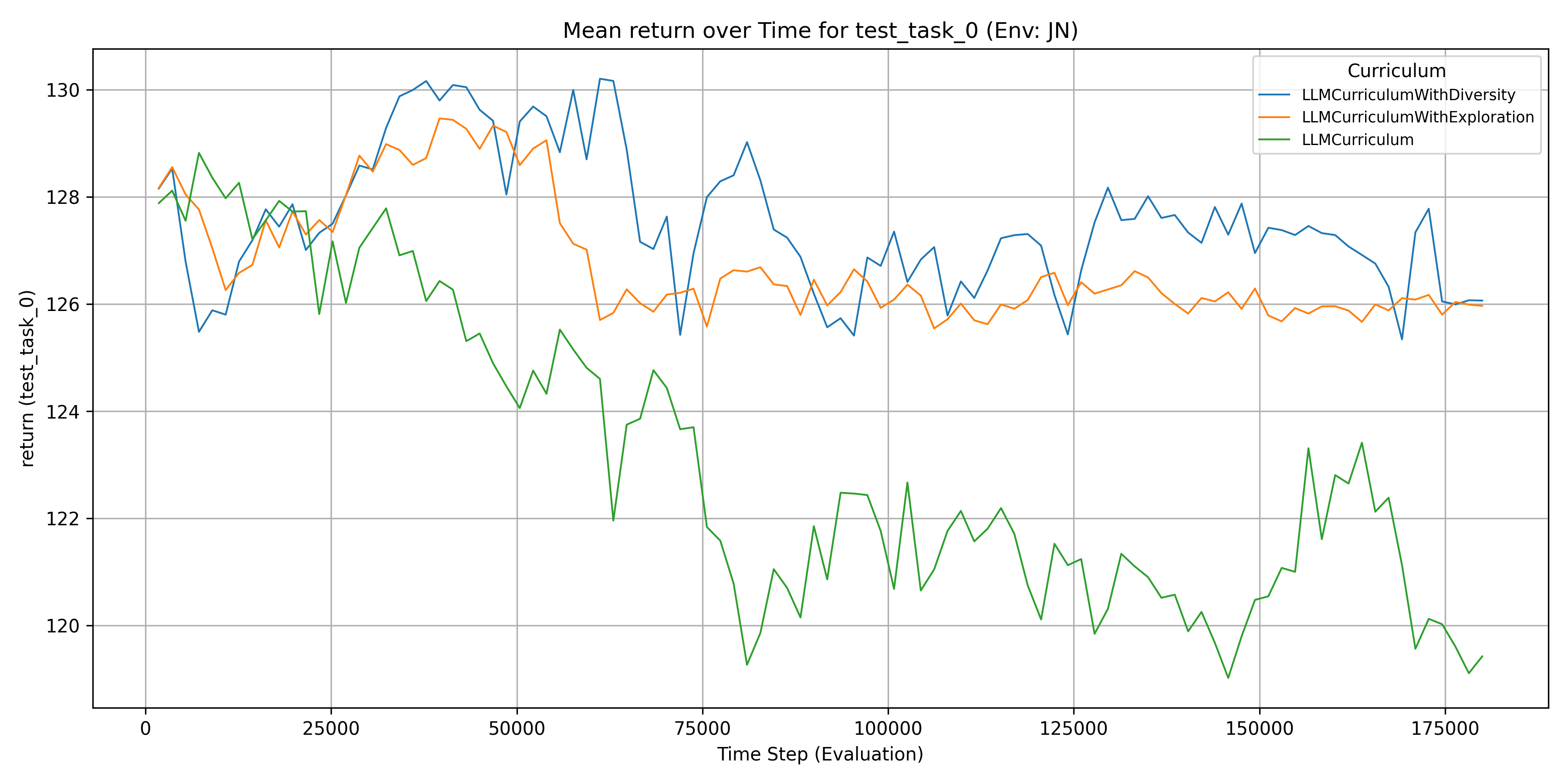}
    \caption{Jinan (3×4), Task 1}
    \label{fig:task0_JN}
  \end{subfigure}
  
  \begin{subfigure}[b]{0.32\textwidth}
    \includegraphics[width=\linewidth]{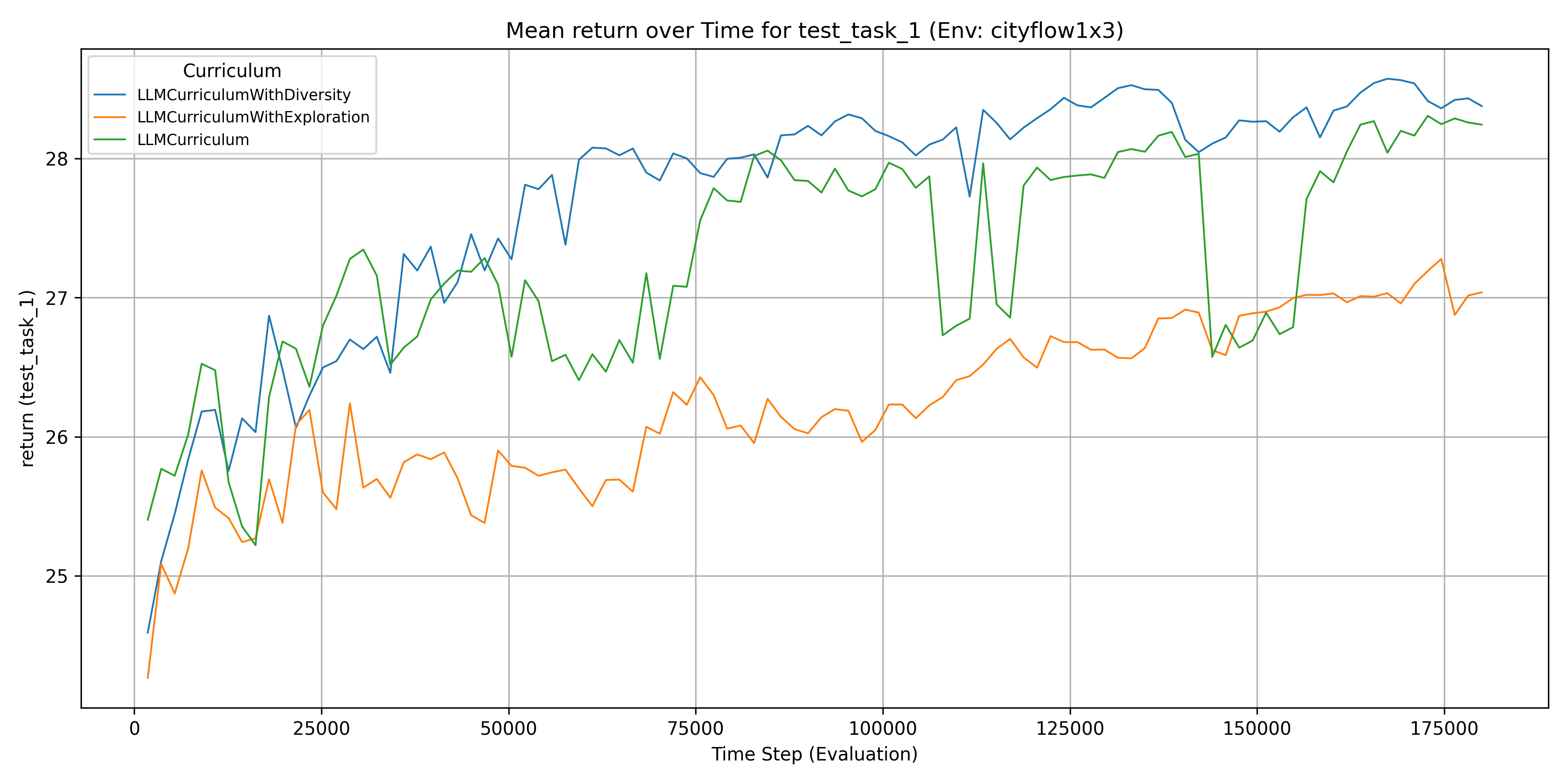}
    \caption{Jinan (1×3), Task 2}
    \label{fig:task1_cityflow1x3}
  \end{subfigure}%
  \begin{subfigure}[b]{0.32\textwidth}
    \includegraphics[width=\linewidth]{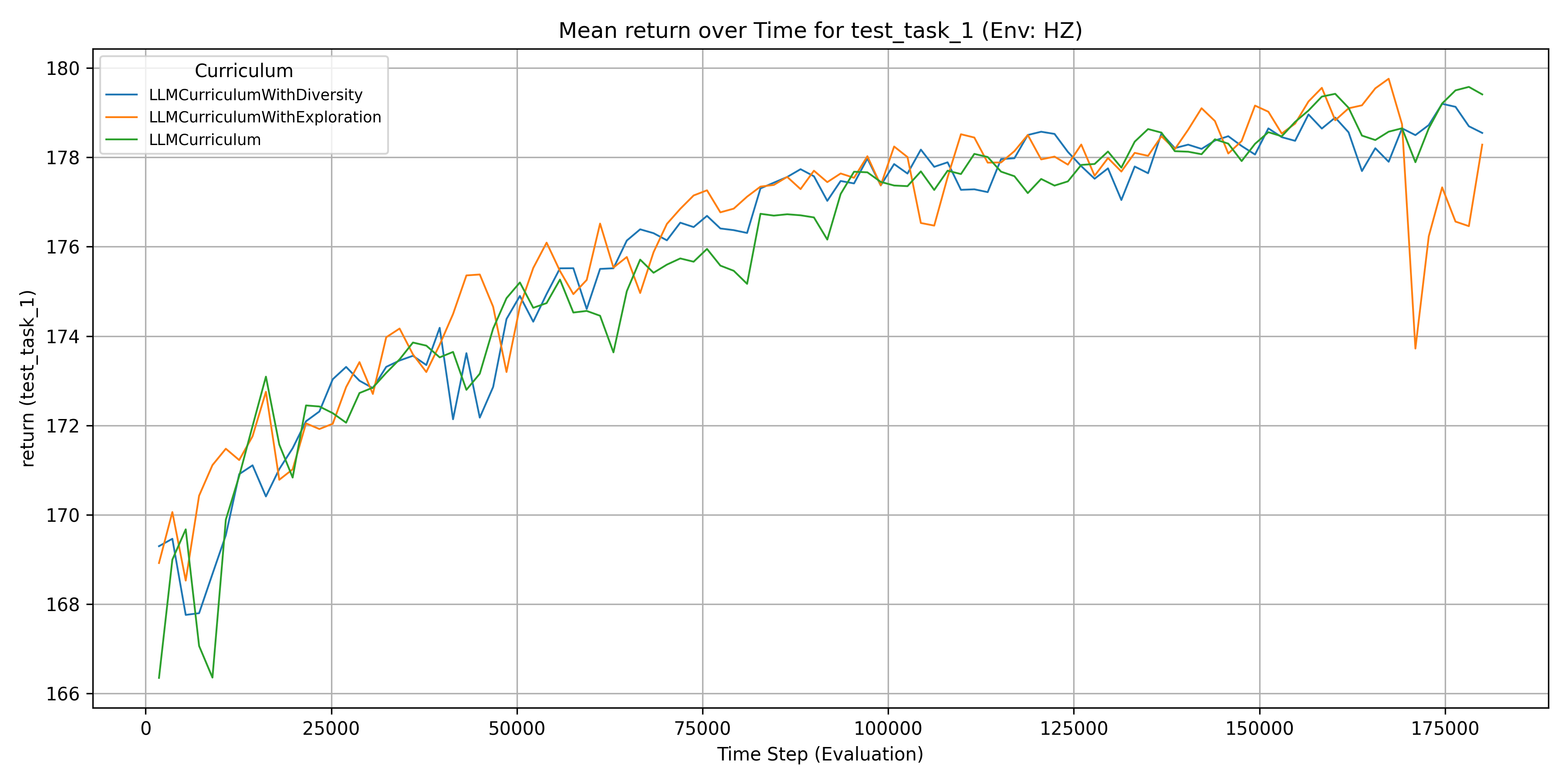}
    \caption{Hangzhou, Task 2}
    \label{fig:task1_HZ}
  \end{subfigure}%
  \begin{subfigure}[b]{0.32\textwidth}
    \includegraphics[width=\linewidth]{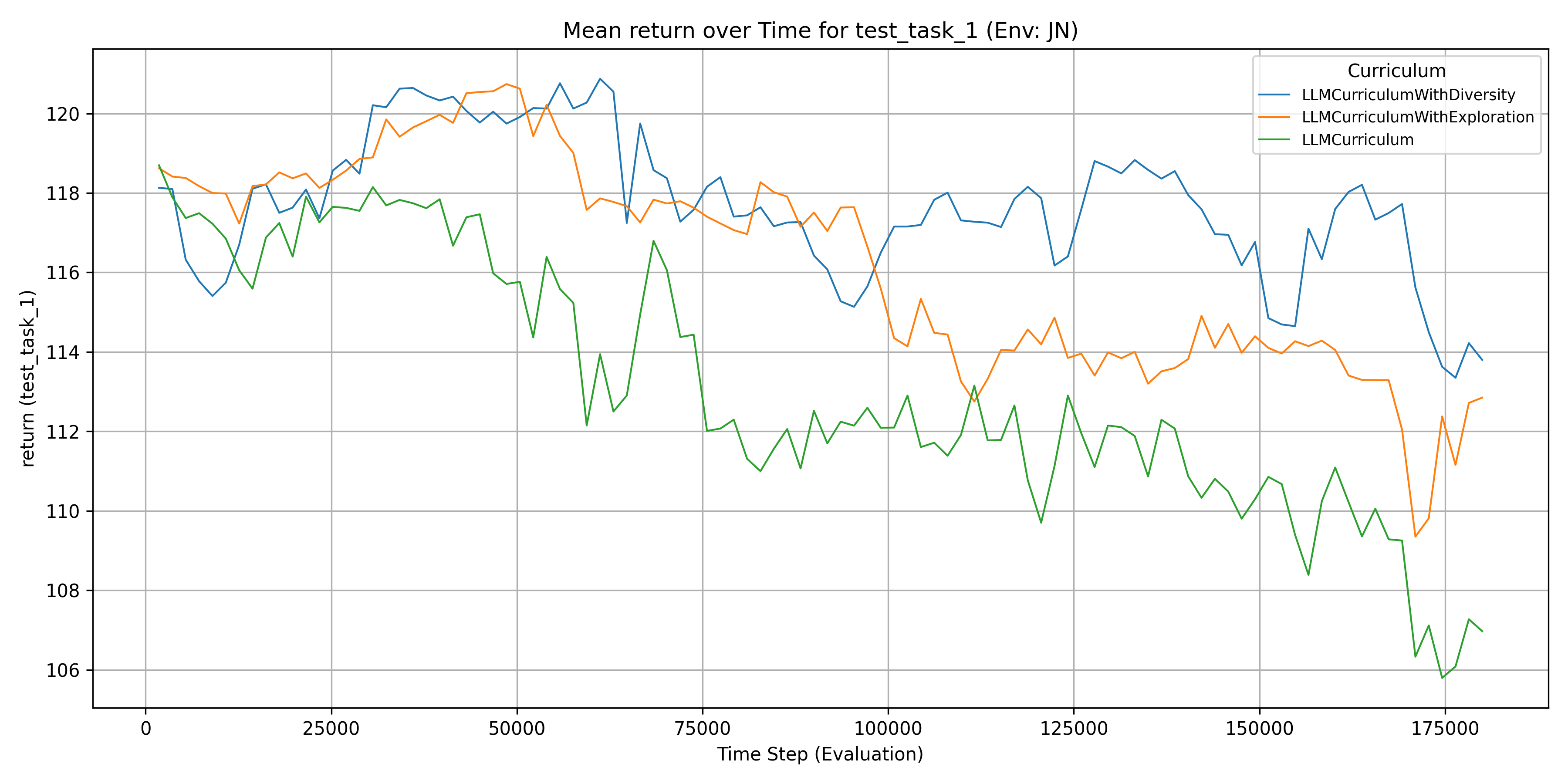}
    \caption{Jinan (3×4), Task 2}
    \label{fig:task1_JN}
  \end{subfigure}
  
  \begin{subfigure}[b]{0.32\textwidth}
    \includegraphics[width=\linewidth]{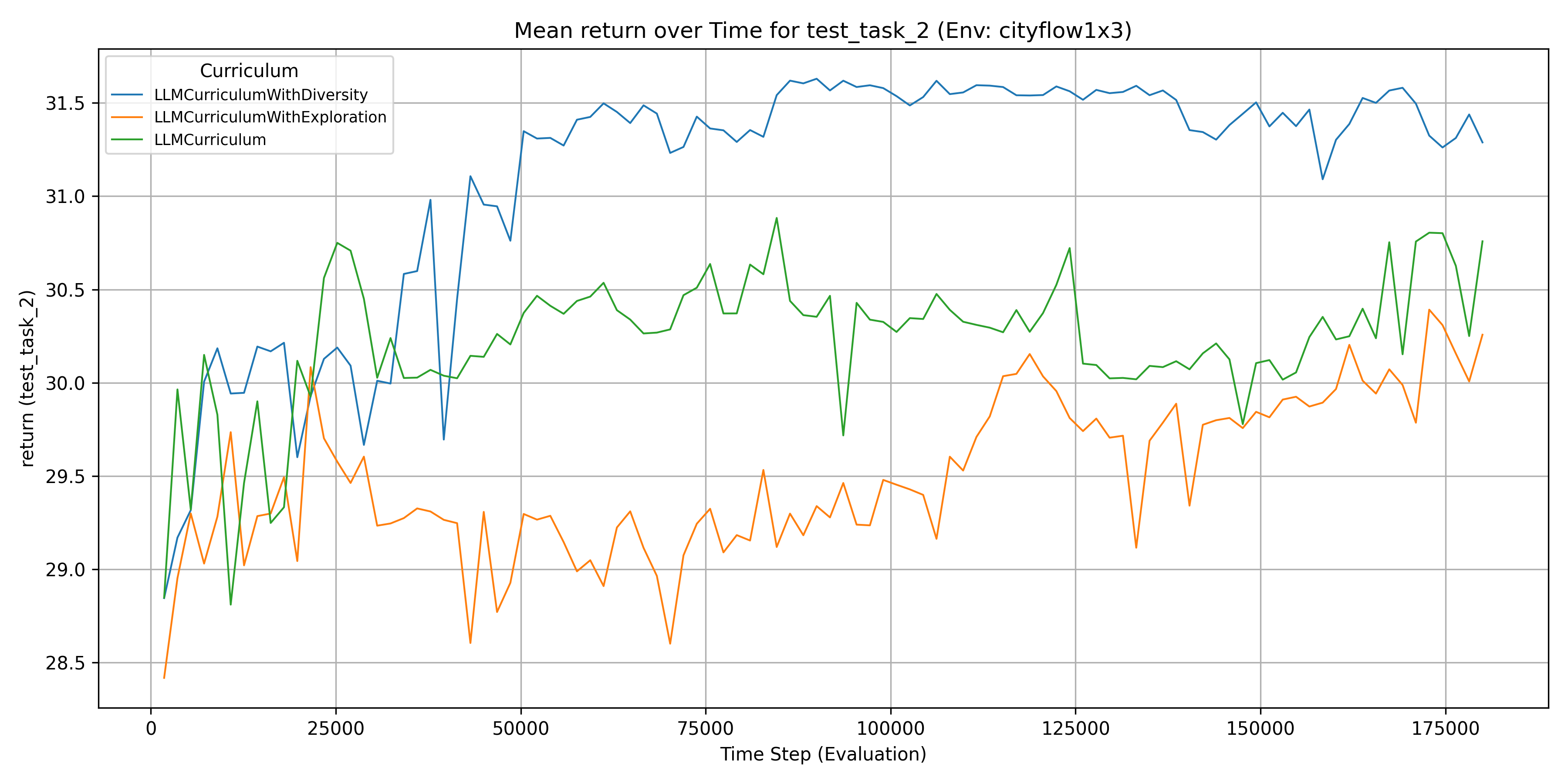}
    \caption{Jinan (1×3), Task 3}
    \label{fig:task2_cityflow1x3}
  \end{subfigure}%
  \begin{subfigure}[b]{0.32\textwidth}
    \includegraphics[width=\linewidth]{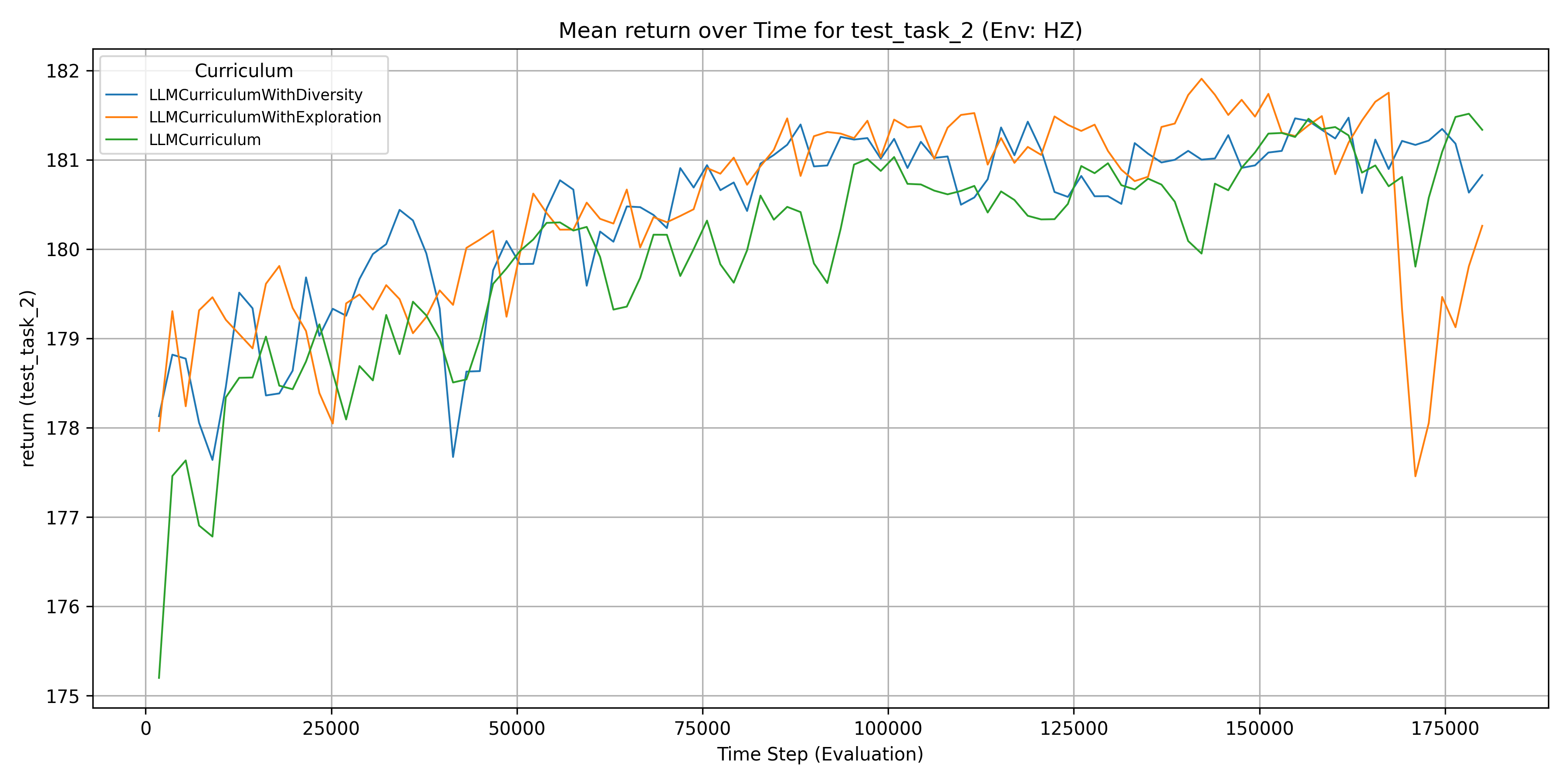}
    \caption{Hangzhou, Task 3}
    \label{fig:task2_HZ}
  \end{subfigure}%
  \begin{subfigure}[b]{0.32\textwidth}
    \includegraphics[width=\linewidth]{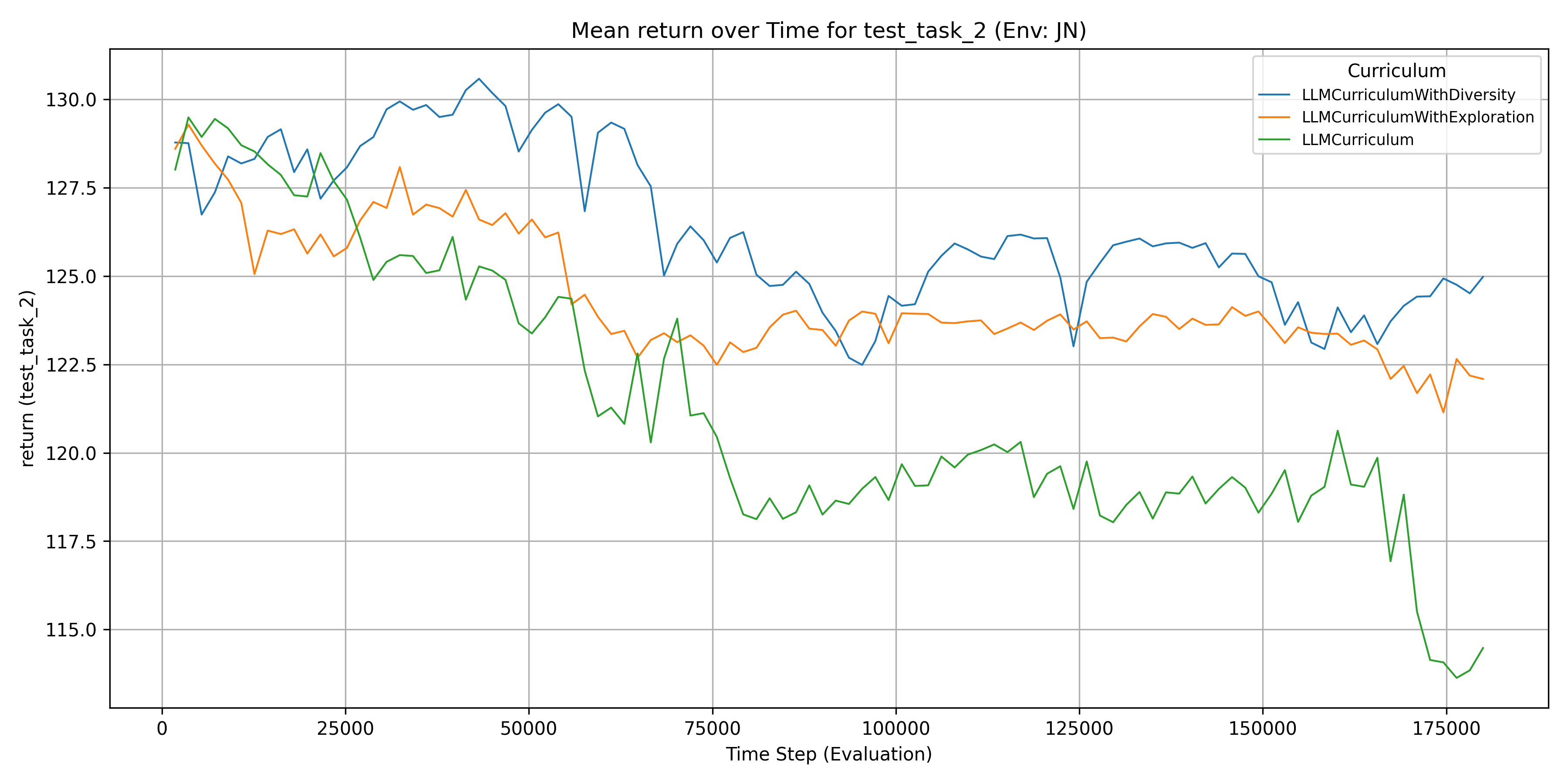}
    \caption{Jinan (3×4), Task 3}
    \label{fig:task2_JN}
  \end{subfigure}
  
  \begin{subfigure}[b]{0.32\textwidth}
    \includegraphics[width=\linewidth]{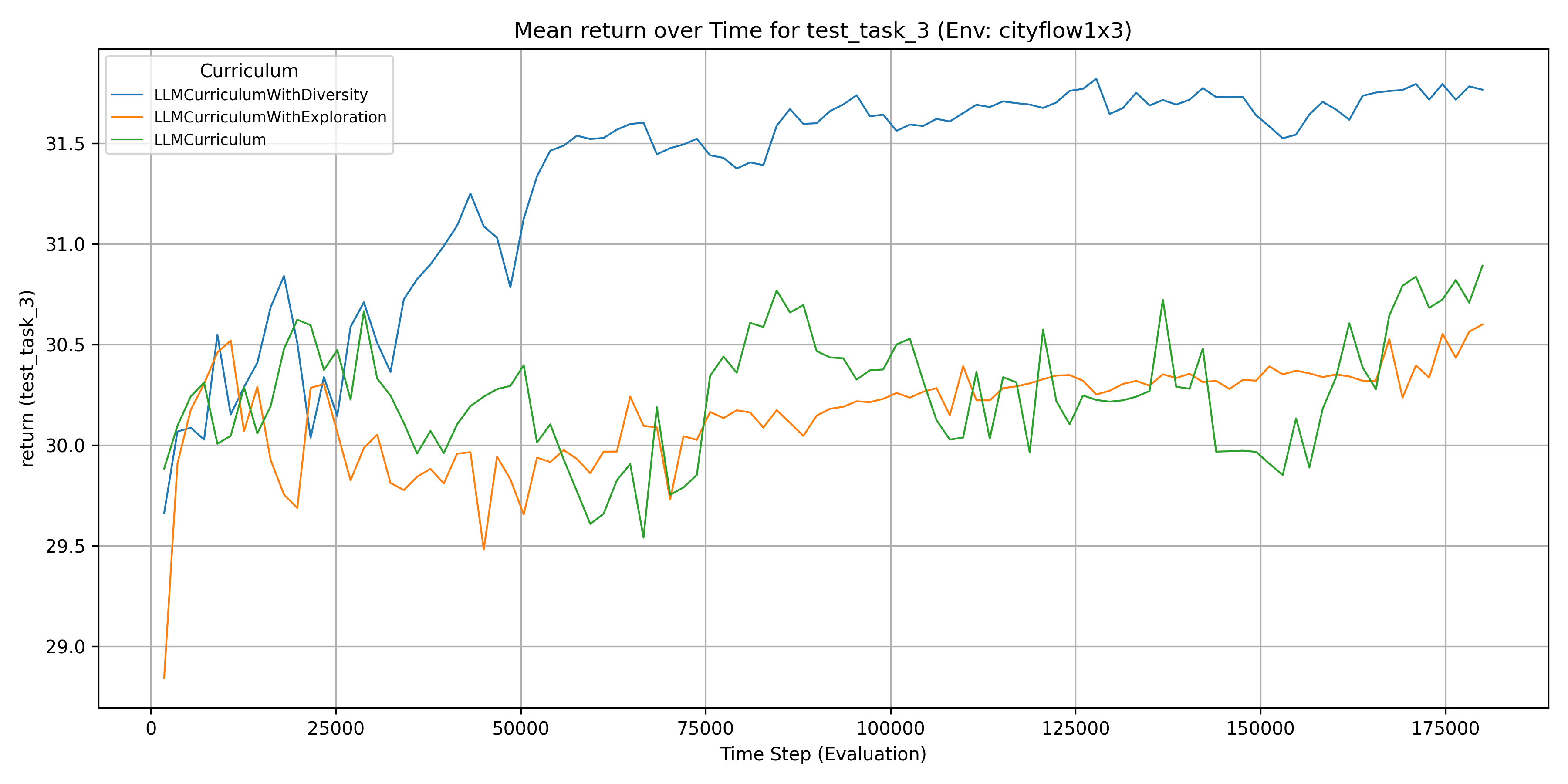}
    \caption{Jinan (1×3), Task 4}
    \label{fig:task3_cityflow1x3}
  \end{subfigure}%
  \begin{subfigure}[b]{0.32\textwidth}
    \includegraphics[width=\linewidth]{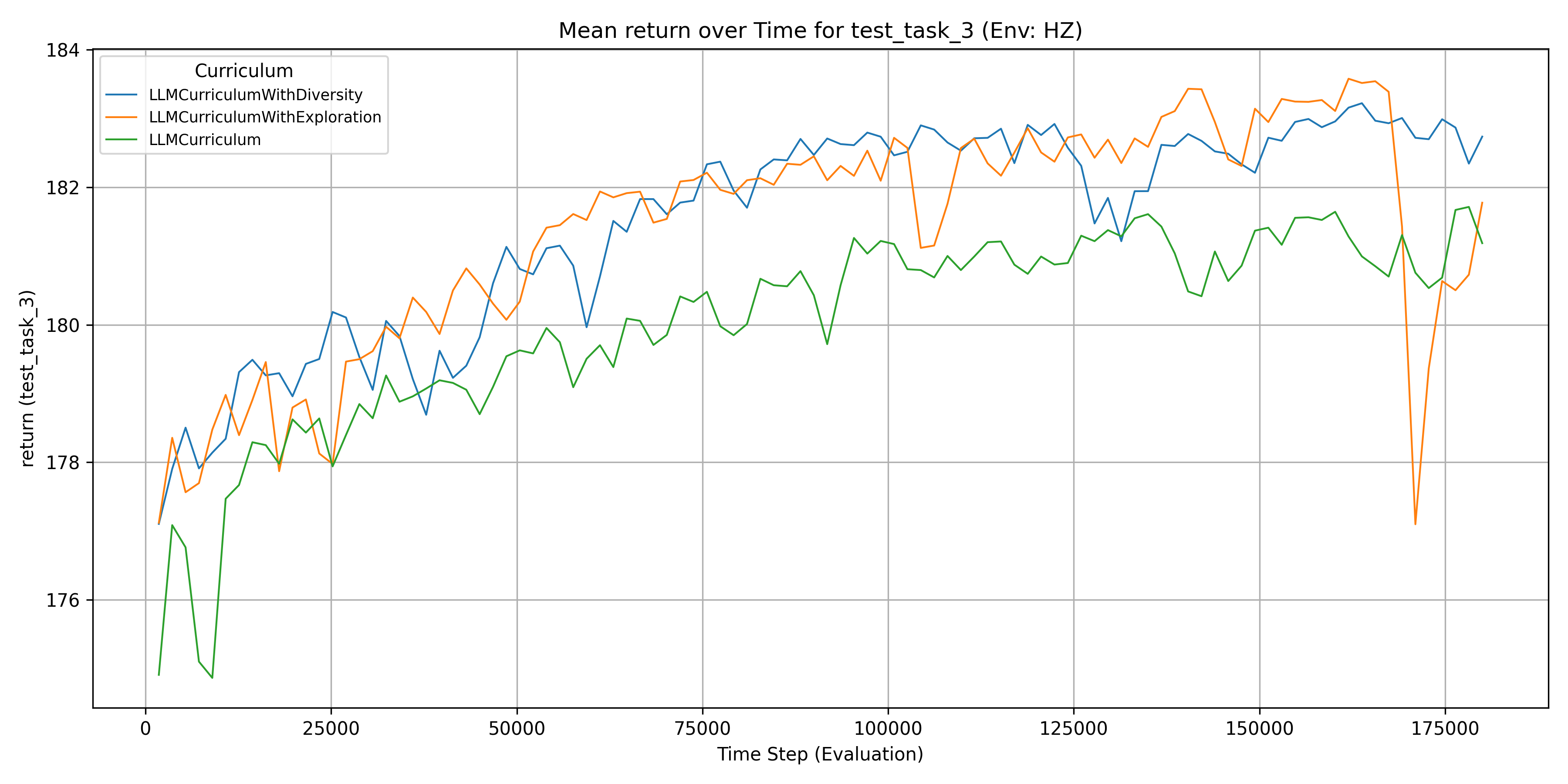}
    \caption{Hangzhou, Task 4}
    \label{fig:task3_HZ}
  \end{subfigure}%
  \begin{subfigure}[b]{0.32\textwidth}
    \includegraphics[width=\linewidth]{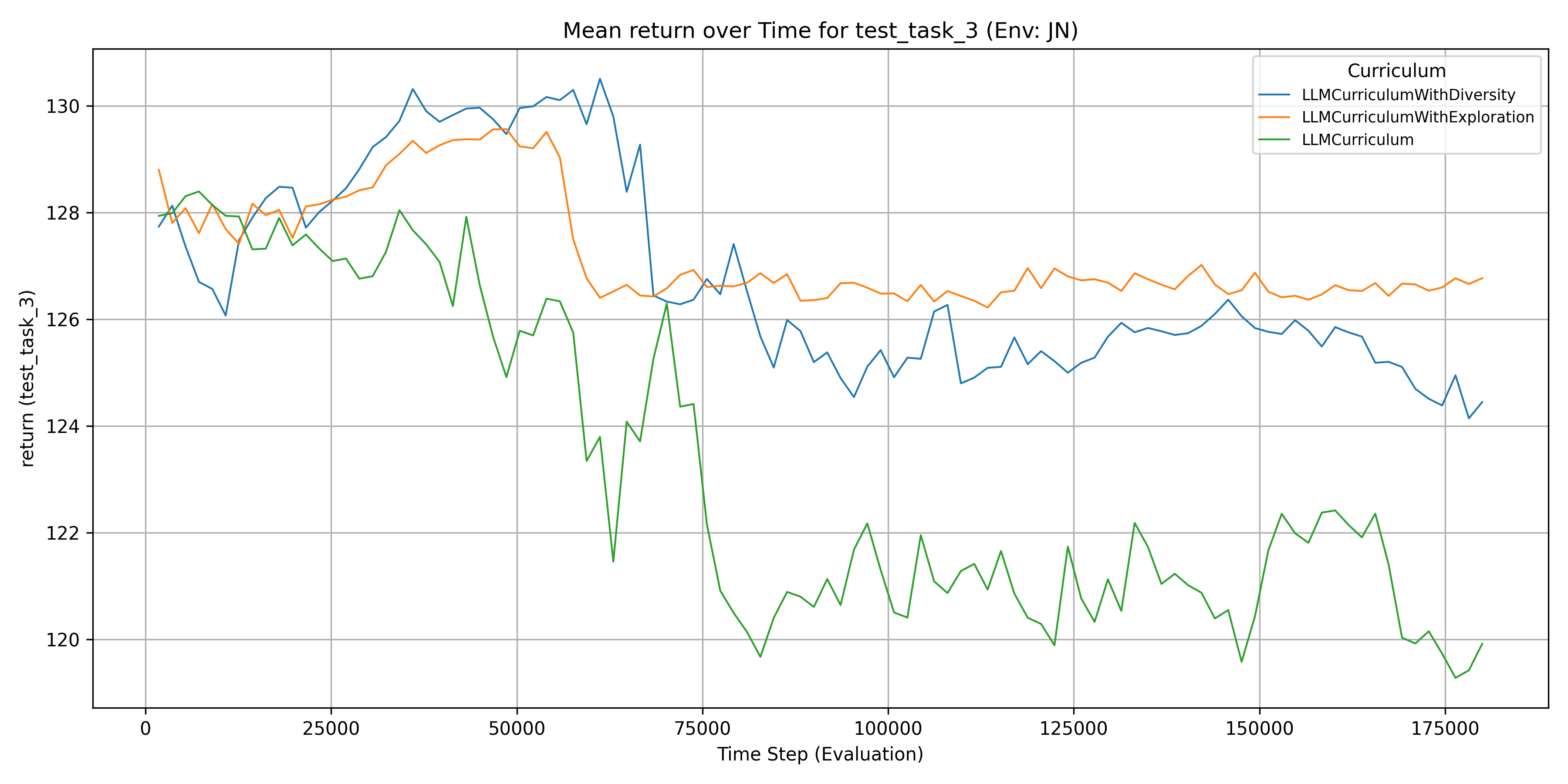}
    \caption{Jinan (3×4), Task 4}
    \label{fig:task3_JN}
  \end{subfigure}
  
  \begin{subfigure}[b]{0.32\textwidth}
    \includegraphics[width=\linewidth]{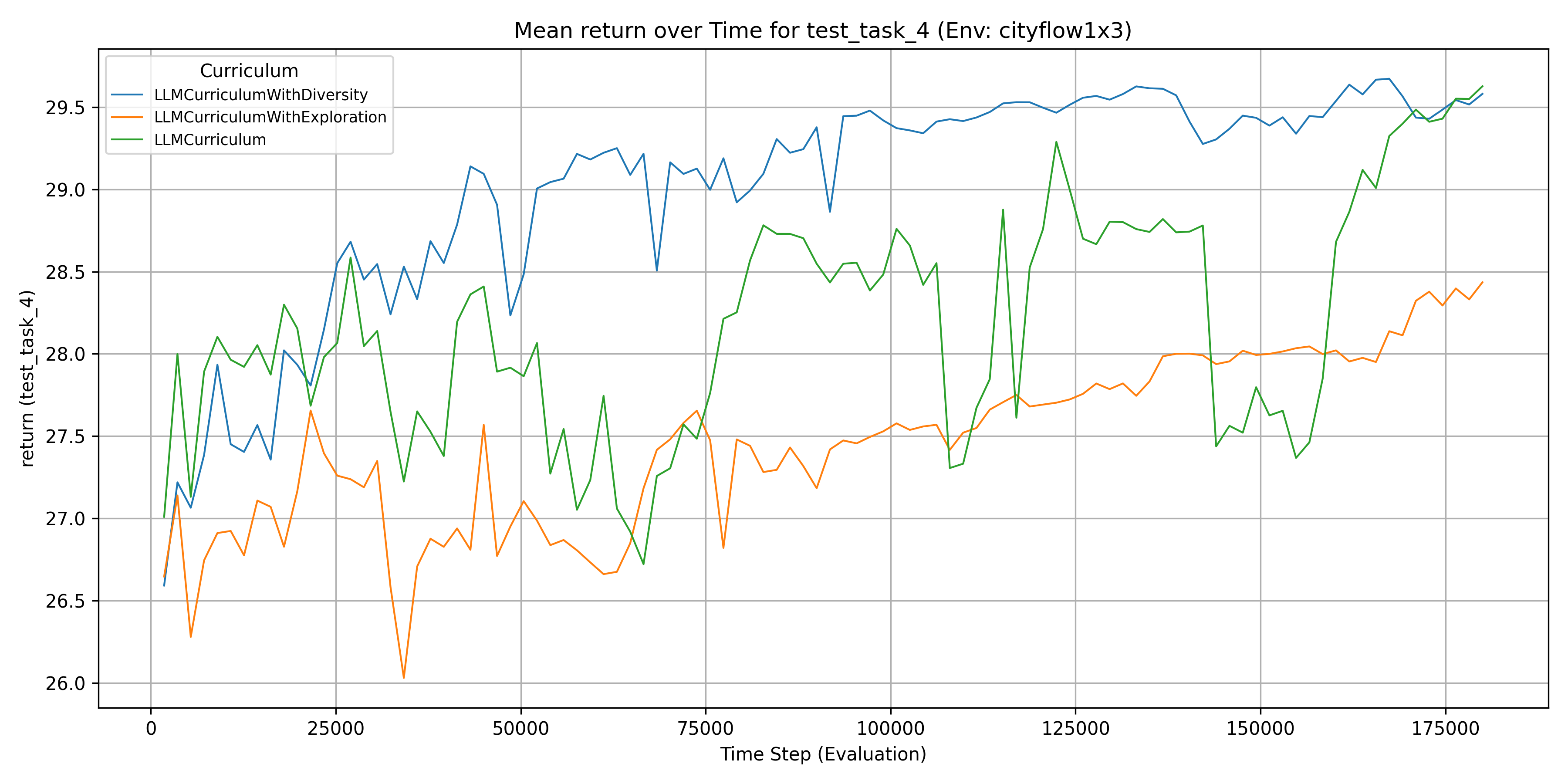}
    \caption{Jinan (1×3), Task 5}
    \label{fig:task4_cityflow1x3}
  \end{subfigure}%
  \begin{subfigure}[b]{0.32\textwidth}
    \includegraphics[width=\linewidth]{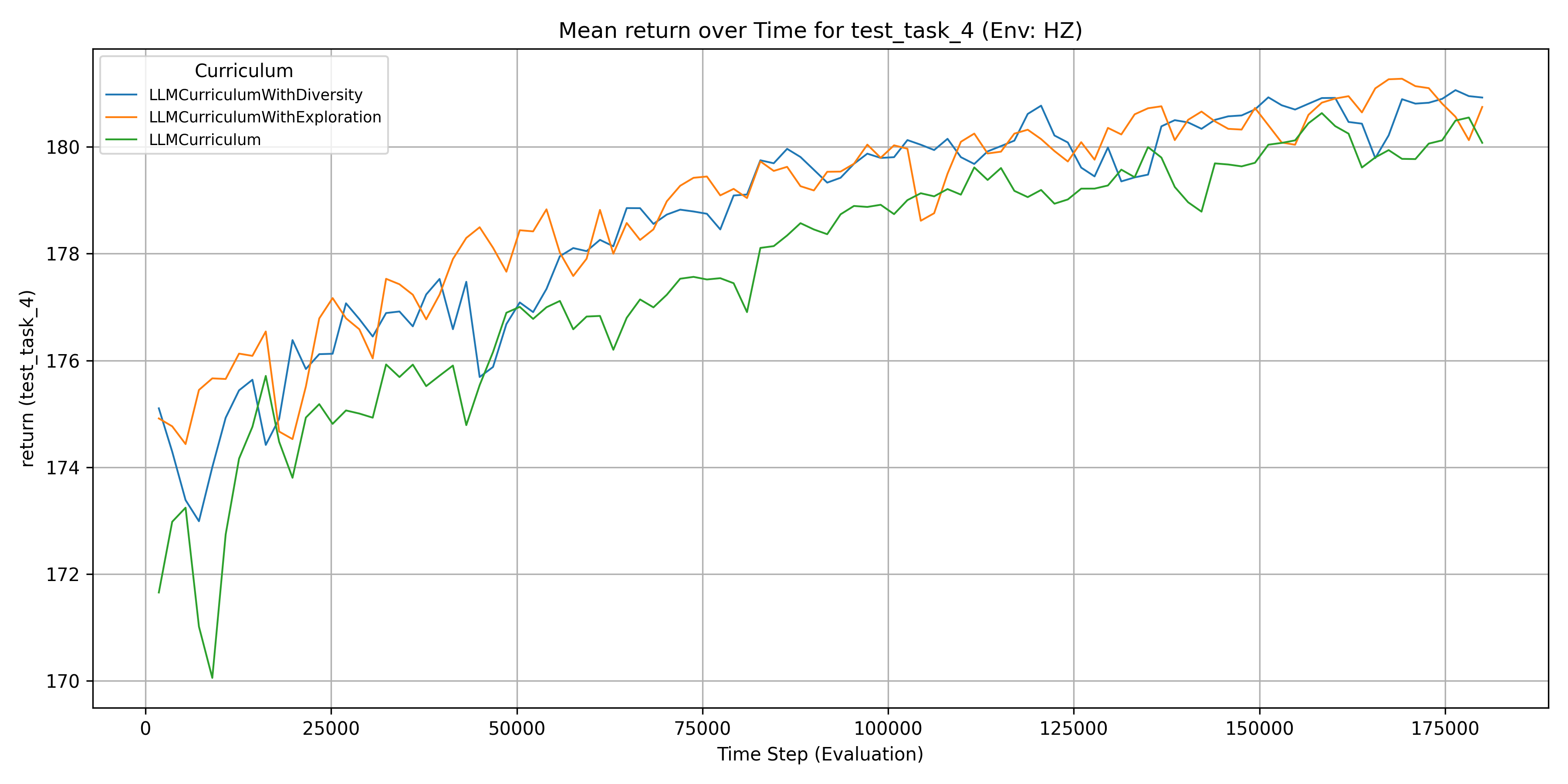}
    \caption{Hangzhou, Task 5}
    \label{fig:task4_HZ}
  \end{subfigure}%
  \begin{subfigure}[b]{0.32\textwidth}
    \includegraphics[width=\linewidth]{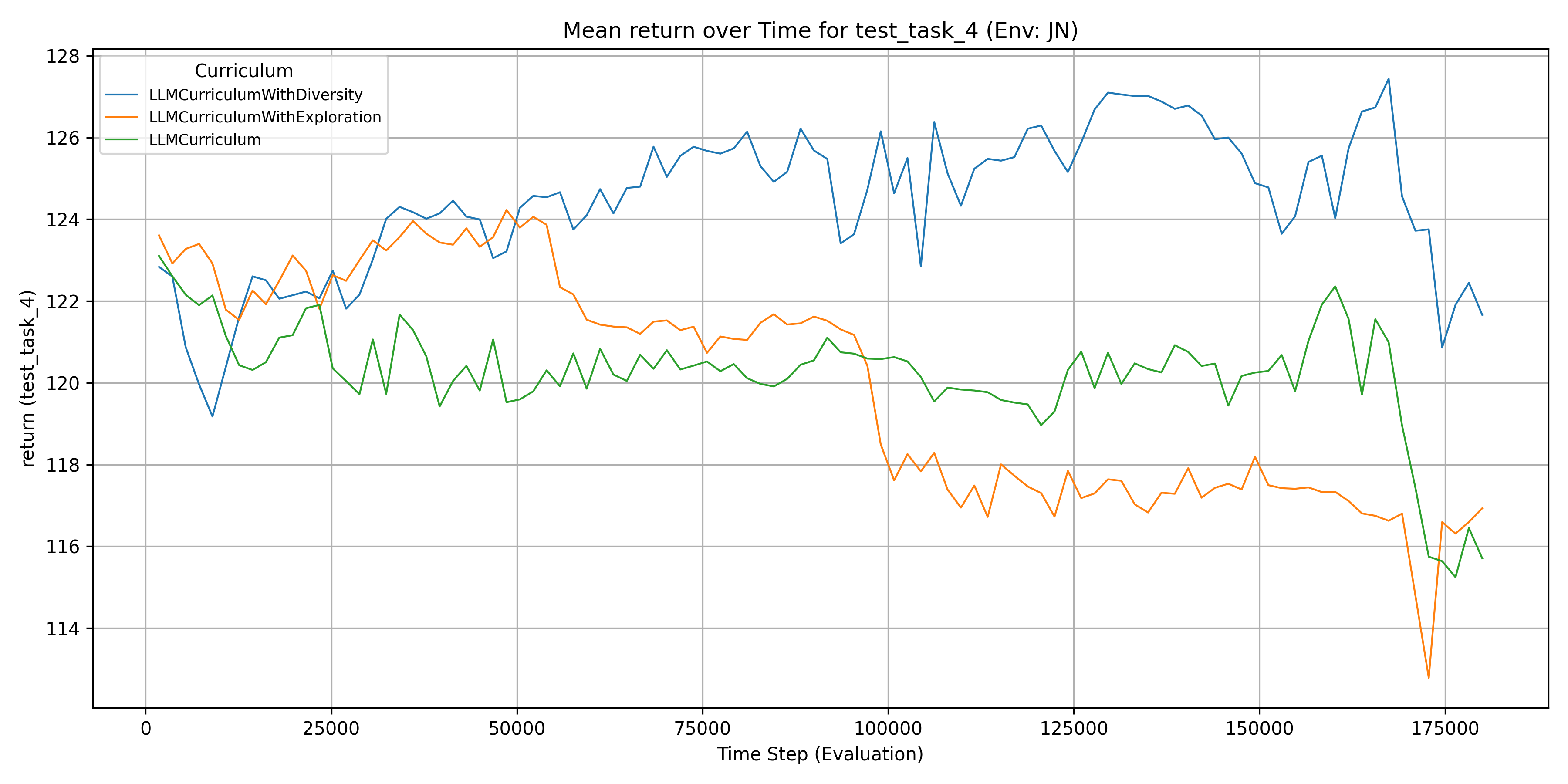}
    \caption{Jinan (3×4), Task 5}
    \label{fig:task4_JN}
  \end{subfigure}
  
  \caption{
  Test return over training timesteps across three datasets (columns) and five held-out test tasks (rows). Each plot shows the performance of three algorithm variants (cMALC-D, cMALC, and cMALC-$\epsilon$) averaged over five random seeds. The left column shows results for the Jinan (1×3) dataset, the middle column for Hangzhou, and the right column for the more challenging Jinan (3×4) dataset.
  }
  \label{fig:diversity_ablation}
\end{figure*}

Figure~\ref{fig:diversity_ablation} shows the mean test return across five held-out test tasks (rows) and three datasets (columns), with each curve averaged over five random seeds. On the easier Jinan \((1 \times 3)\) dataset (left column), cMALC-D yields clear performance improvements. This is especially evident in test tasks 1, 3, and 4 (first, third, and fourth rows), where cMALC-D consistently achieves a test return around 31, outperforming the other variants by approximately two points on average.

On the Hangzhou dataset (middle column), performance across all three variants is more similar, but cMALC-D still shows consistent gains. Across all settings, the inclusion of the diversity mechanism improves sample efficiency and accelerates convergence. However, cMALC-\(\epsilon\) exhibits greater instability. In particular, test tasks 1 through 4 (first four rows) show that cMALC-\(\epsilon\) often suffers sharp drops in performance around 170,000 timesteps, sometimes losing much of the progress previously made.

Performance is uniformly lower on the most challenging Jinan \((3 \times 4)\) dataset (right column). We attribute this to the increased traffic density, as small changes in the context can cause substantial environment changes. Nevertheless, cMALC-D remains the top-performing variant. Notably, the baseline cMALC frequently shows a decline in test reward over time (particularly visible in test tasks 1-4), which suggests mode collapse and poor generalization to test contexts, further highlighting the need for context diversity during training.

\subsection{What kinds of contexts are generated?}
To better understand how curricula evolve, we visualize context trajectories using heatmaps in Figures~\ref{fig: individual_feature_values_cityflow1x3}, \ref{fig: individual_feature_values_HZ}, and \ref{fig: individual_feature_values_JN}. Across all three environments, we observe that context features tend to evolve gradually over time rather than undergoing abrupt shifts. This gradual progression facilitates a smoother learning process for the agents and supports our finding that cMALC-D outperforms Domain Randomization due to a more controlled approach of generating diverse contexts.

We also observe that in the curriculum for the JN \(1 \times 3\) environment, multiple context features are frequently updated simultaneously. For instance, around episode 125, changes occur simultaneously in \texttt{maxSpeed} and \texttt{minGap}, along with adjustments to vehicle \texttt{length} and \texttt{width}. A similar cluster of updates appears around episode 375, involving both \texttt{maxNegAcc} and \texttt{usualNegAcc}, and in the HZ environment near episode 450.

This raises an interesting question: \textit{do LLMs internally form or leverage semantic relationships between features, such as coupling speed with spacing, or braking with acceleration, to propose more coherent or pedagogically aligned context transitions? } To answer this question, we construct correlation matrices between features in the cMALC-D generated curriculum in Figures \ref{fig:cityflow1x3_correlation}, \ref{fig:HZ_correlation}, \ref{fig:JN_correlation}.

In the JN \(1 \times 3\) environment, which contains fewer vehicles and simpler dynamics, the LLM identifies and exploits meaningful semantic patterns. For example, \texttt{maxNegAcc} (maximum negative acceleration or braking) has a weak correlation with other features, suggesting that the model implicitly recognizes the limited need for braking behaviors due to the sparsity of interactions in a small-scale scenario. In the HZ environment, vehicle width shows a similarly low correlation, despite the model not being given lane dimensions. This suggests the LLM infers structural properties of the environment, like tight lane-to-vehicle width ratios, and develops curricula that do not depend on this feature, purely from context patterns. Such reasoning allows it to generate more coherent and pedagogically effective curricula without explicit environmental parameters.

On the other hand, the LLM also captures strong correlations between semantically coupled features. For example, in the JN \(3 \times 4\) environment (Figure~\ref{fig:cityflow1x3_correlation}), the correlation between \texttt{maxSpeed} and \texttt{minGap} is 0.71, reflecting the LLM’s implicit understanding that higher speeds require larger spacing between vehicles to ensure safety. Similarly, in the HZ environment (Figure~\ref{fig:HZ_correlation}), \texttt{usualPosAcc} and \texttt{maxPosAcc} are correlated with a value of 0.83, indicating the LLM’s awareness that general acceleration behavior is often constrained by maximum performance limits.


\section{Conclusion and Future Work}
In this paper, we develop cMALC-D, an LLM-based curriculum learning algorithm for contextual MARL. Our method leverages the reasoning capabilities of LLMs to generate semantically meaningful curricula. We also introduce a novel diversity-based mechanism based on task arithmetic from continual learning to encourage exploration in the context space and avoid mode collapse. Our experiments on three real-world traffic environments show that cMALC-D enhances MARL policy generalization and sample efficiency over a variety of environment configurations.

While our formulation is based on modifying the transition function by altering context features, there are other methods to model different environment configurations. For example, in the traffic signal control setting, sensor malfunctions can cause agents to return noisy data \citep{yang2024mallightinfluenceawarecoordinatedtraffic}. One possible way to include this is to have the LLM generate a more general context, such as a noisy environment configuration parameterized by a probability distribution. Another avenue for future work is explicitly encoding semantic feature relationships to augment self-paced curricula.

\bibliography{example_paper}
\bibliographystyle{icml2025}

\newpage
\appendix
\onecolumn
\section{Additional Experimental Details}
\label{sec: additional experiments details}

\subsection{Dataset Details}
\label{subsec: dataset details}
We use the Cityflow environment with three publicly available datasets derived from real-world data. We use the Jinan \((1 \times 3)\) dataset from PressLight \citep{presslight2019} and the Jinan \((3 \times 4)\) and Hangzhou \((4 \times 4)\) datasets from CoLight \citep{colight2019}. Each dataset contains a \verb|roadnet.json| file that gives the topology of the road network and a \verb|flow.json| file that gives the vehicle routes and flow patterns. A more detailed list of dataset statistics can be found in Table \ref{tab:data_stats}.

\subsection{Environment Details} 
\label{subsec: environment details}
\begin{figure}[h]
  \centering
  \includegraphics[width=1\linewidth]{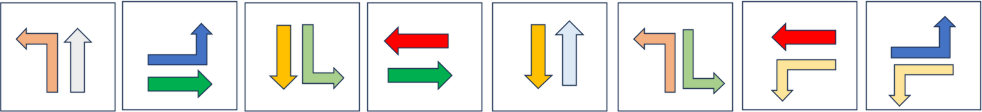}
  \caption{Illustration of all 8 phases for each intersection. Phases are designed so that only two lanes can be active at any time.}
  \label{fig: traffic phases}
\end{figure}

In our experiments, each agent corresponds to a traffic light controlling an intersection. Therefore, the Jinan environments have 3 and 12 agents, respectively, while the Hangzhou environment has 16 agents. Each traffic light agent can choose from 8 possible actions, where each action represents a specific signal phase. A phase refers to a predefined combination of green and red signals assigned to incoming lanes, typically allowing certain directions of traffic (e.g., straight-through, left-turn) while restricting others. Figure \ref{fig: traffic phases} illustrates each phase. The observation for each intersection is composed of information about all ingoing and outgoing cars in each lane, such as their position, velocity, and lane number. Each agent has partial observability and cannot use information that is outside the local observation to make decisions. At time \(t\), the global state \(s_t\) is the concatenation of all local observations. The global reward is defined as \(\lambda_fT_f + \lambda_wT_w\), where \(T_f\) is the total time vehicles are moving, \(T_w\) is the total time vehicles are waiting in traffic, and \(\lambda_f, \lambda_w\) are hyperparameters. In practice, we set \(\lambda_f = 0.033\) and \(\lambda_w = 0\).

We evaluate our traffic control policies using several real-world metrics that reflect their effectiveness. Wait time measures the total duration that vehicles remain stationary (i.e., with zero speed) during their journeys. Delay time captures the additional time each vehicle takes to reach its destination compared to the ideal travel time in the absence of any traffic signals or congestion. Finally, we calculate the throughput, which is the total number of vehicles that complete their routes before the end of the episode.

\begin{table}[h]
\centering
\begin{tabular}{lccccccc}
\toprule
\textbf{Dataset} & \textbf{\# Intersections} & \textbf{Mean} & \textbf{Std} & \textbf{Max} & \textbf{Min} & \textbf{\# Lanes} & \textbf{Time Steps}\\
\midrule
Hangzhou & 16 & 526.63 & 86.70 & 676 & 256 & 3 & 3600 \\
Jinan \((3 \times 4)\) & 12 & 250.70 & 38.21 & 335 & 208 & 3 & 3600\\
Jinan \((1 \times 3)\) & 3 & 278.23 & 18.47 & 322 & 229 & 3 & 3600\\
\bottomrule
\end{tabular}
\caption{Data statistics of real-world traffic datasets. Arrival rates are reported vehicles per 300 seconds.}
\label{tab:data_stats}
\end{table}

\subsection{Context Parameterization Details}
\label{subsec: context parameterization}
\begin{table}[H]
\centering
\begin{tabular}{lll}
\toprule
\textbf{Parameter} & \textbf{Description} & \textbf{Range} \\
\midrule
\texttt{length} & Length of each car & $1$–$10$ m \\
\texttt{width} & Width of each car & $1$–$5$ m \\
\texttt{maxPosAcc} & Max acceleration when speeding up & $0.5$–$5$ m/s\textsuperscript{2} \\
\texttt{maxNegAcc} & Max deceleration when braking & $0.5$–$5$ m/s\textsuperscript{2} \\
\texttt{usualPosAcc} & Default acceleration when speeding up & $1$–$5$ m/s\textsuperscript{2} \\
\texttt{usualNegAcc} & Default deceleration when braking & $1$–$5$ m/s\textsuperscript{2} \\
\texttt{minGap} & Minimum gap between cars & $1$–$10$ m \\
\texttt{maxSpeed} & Maximum speed a car can travel & $3$–$15$ m/s \\
\texttt{headwayTime} & Desired time to reach the vehicle in front & $1$–$5$ s (int) \\
\bottomrule
\end{tabular}
\caption{Context parameters used for curriculum learning and their specified ranges.}
\label{tab:context_params}
\end{table}

\subsection{Hyperparameters}
\label{subsec: Hyperparameters}
The MAPPO algorithm is adapted from the ePYMARL library \citep{papoudakis2021benchmarking}. We provide a full list of algorithmic environment hyperparameters in Table \ref{tab: mappo hyperparameters}. We also use an Adam optimizer with a learning rate of $0.003$ to train the cost estimator. We also list the hyperparameters of cMALC-D and the parameters used for the LLM in Tables \ref{tab: hyperparameters for cMALC-D} and \ref{tab: LLM Hyperparameters} respectively.

\begin{table}[h]
\centering
\caption{Hyperparameter for MAPPO}
\label{tab: mappo hyperparameters}
\begin{tabular}{lcccc}
\hline
\textbf{MAPPO Hyperparameters} & \text{} \\ \hline
Eps Clip & 0.2\\
Epsilon Anneal Time & 180000 \\
Learning Rate (lr) & 0.0003 \\
Hidden Dim & 128 \\
Mini Epochs & 4 \\
Entropy Coef & 0.001 \\
Target Update Interval & 0.01 \\
Batch Size & 1 \\
Buffer Size & 10 \\
\hline
\end{tabular}
\end{table}

\begin{table}[h]
\centering
\caption{cMALC-D Hyperparameters}
\begin{tabular}{lcccc}
\hline
\textbf{Hyperparameter} & \text{} \\ \hline
$A$ & MAPPO\\
$M$ & Qwen2.5-7B-Instruct-AWQ \\
$\alpha$ & 0.5 \\
$w$ & 3 \\
$\delta$ & 0.1 \\
$k$ & 3 \\
\hline
\end{tabular}
\label{tab: hyperparameters for cMALC-D}
\end{table}

\begin{table}[h]
\centering
\caption{LLM Hyperparameters}
\begin{tabular}{lcccc}
\hline
\textbf{Hyperparameter} & \text{} \\ \hline
Temperature & 0.7\\
Top-p & 0.9 \\
Max New Tokens & 400 \\
\hline
\end{tabular}
\label{tab: LLM Hyperparameters}
\end{table}

\section{Further Results}
\label{sec: Further Results}
\begin{figure}[H]
    \centering
    \includegraphics[width=\linewidth]{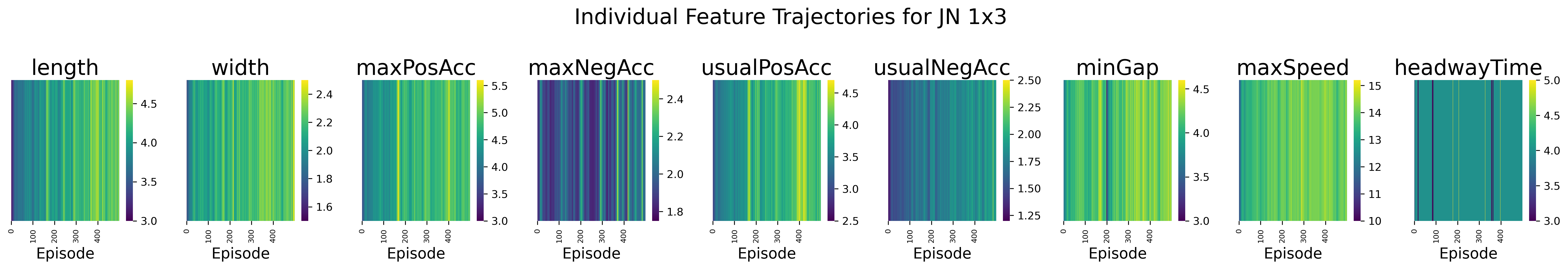}
    \caption{Individual feature values over time for cMALC-D for JN \(1 \times 3\).}
    \label{fig: individual_feature_values_cityflow1x3}
\end{figure}

\begin{figure}[H]
    \centering
    \includegraphics[width=\linewidth]{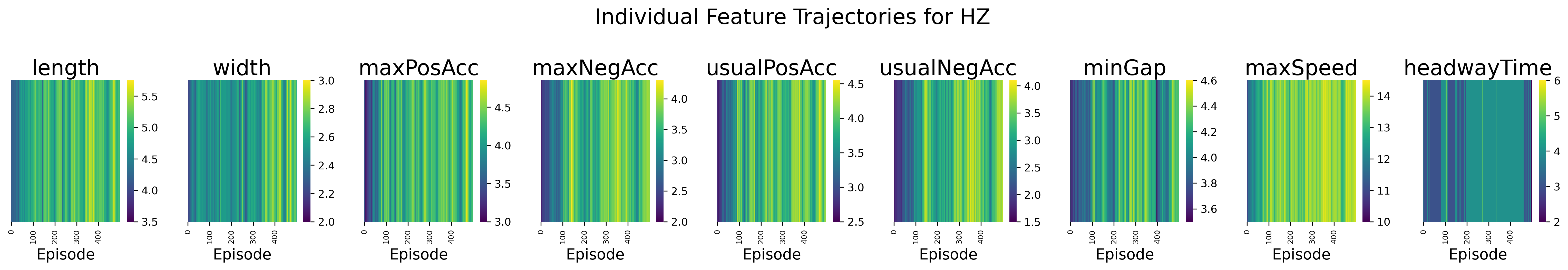}
    \caption{Individual feature values over time for cMALC-D for HZ.}
    \label{fig: individual_feature_values_HZ}
\end{figure}

\begin{figure}[H]
    \centering
    \includegraphics[width=\linewidth]{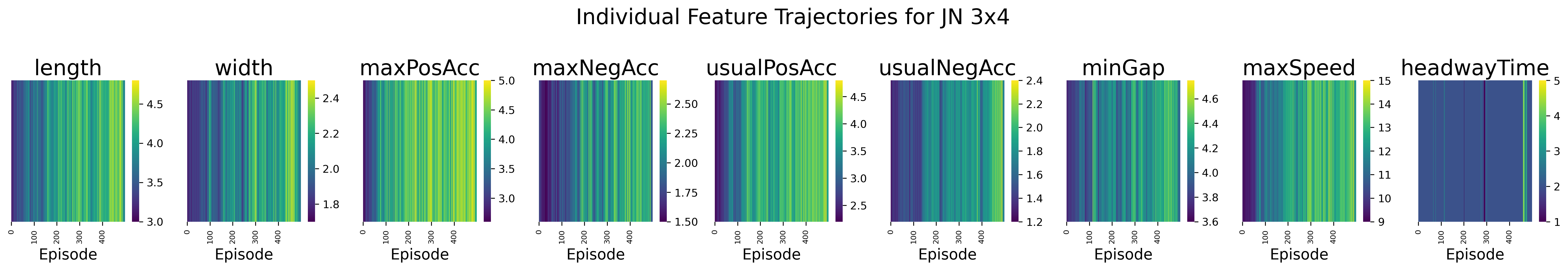}
    \caption{Individual feature values over time for cMALC-D for JN \(3 \times 4\).}
    \label{fig: individual_feature_values_JN}
\end{figure}

\begin{figure}[H]
    \centering
    \includegraphics[width=0.5\linewidth]{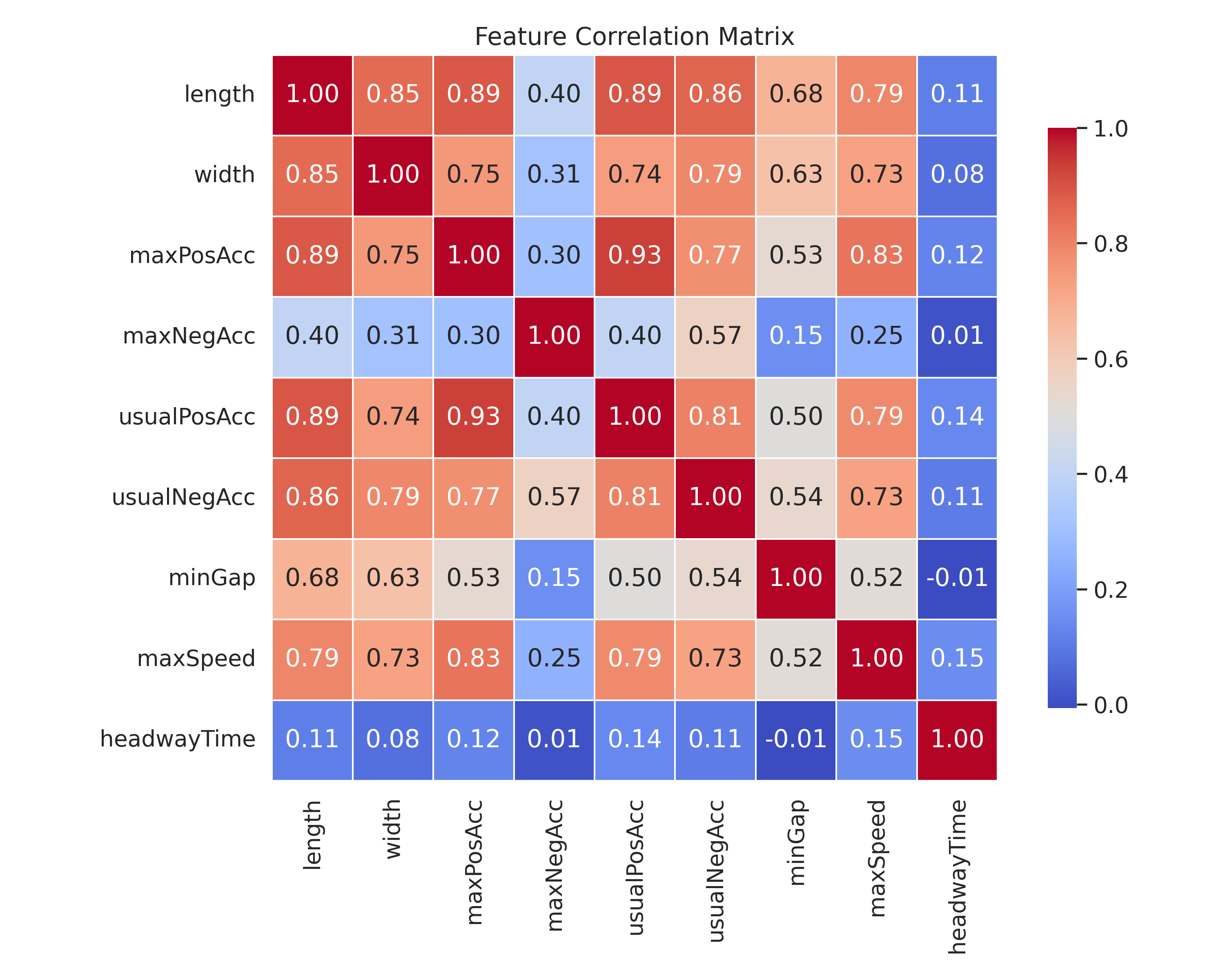}
    \caption{Correlation matrix for cMALC-D generated curriculum on the JN \(1 \times 3\) environment.}
    \label{fig:cityflow1x3_correlation}
\end{figure}

\begin{figure}[H]
    \centering
    \includegraphics[width=0.5\linewidth]{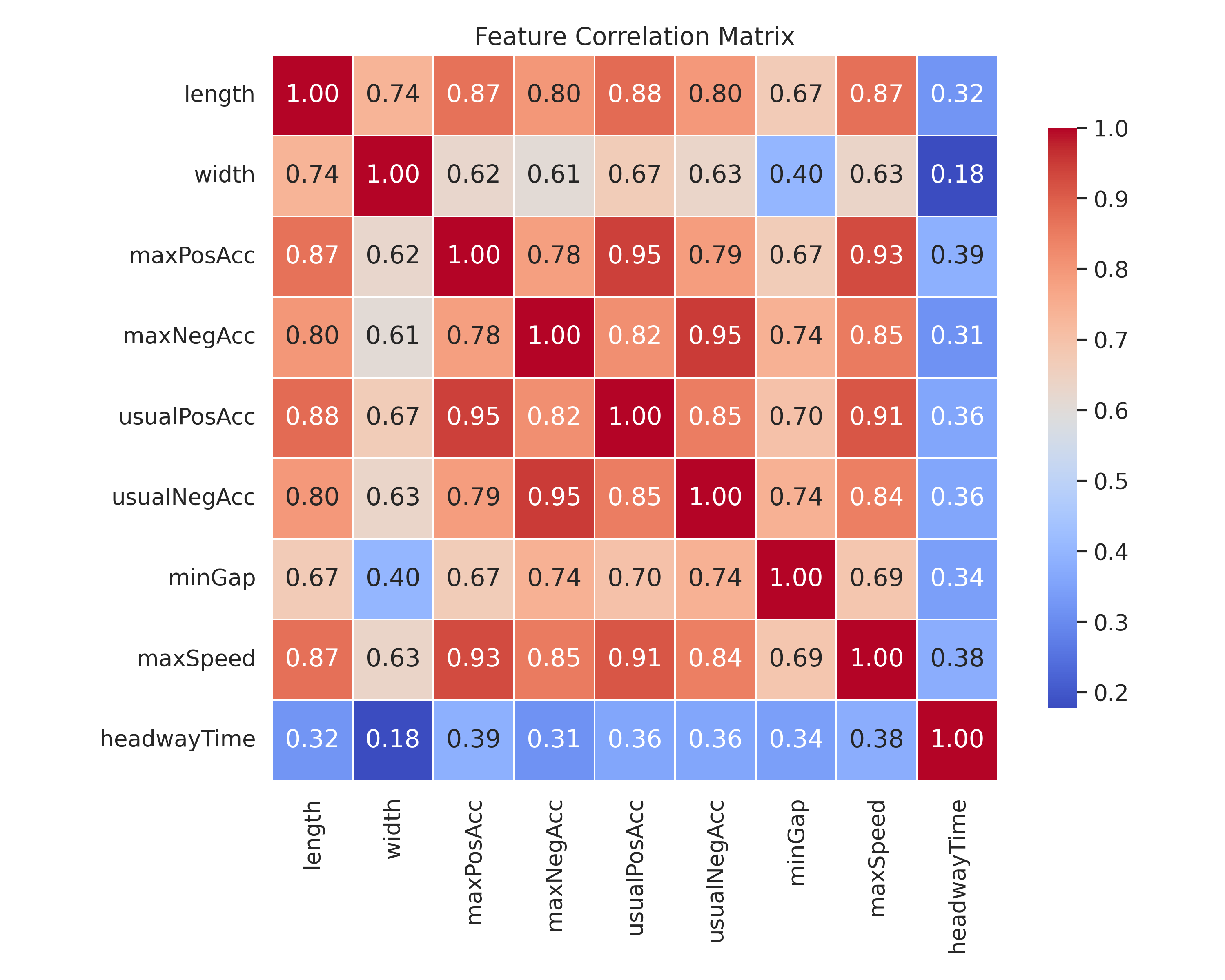}
    \caption{Correlation matrix for cMALC-D generated curriculum on the HZ environment.}
    \label{fig:HZ_correlation}
\end{figure}

\begin{figure}[H]
    \centering
    \includegraphics[width=0.5\linewidth]{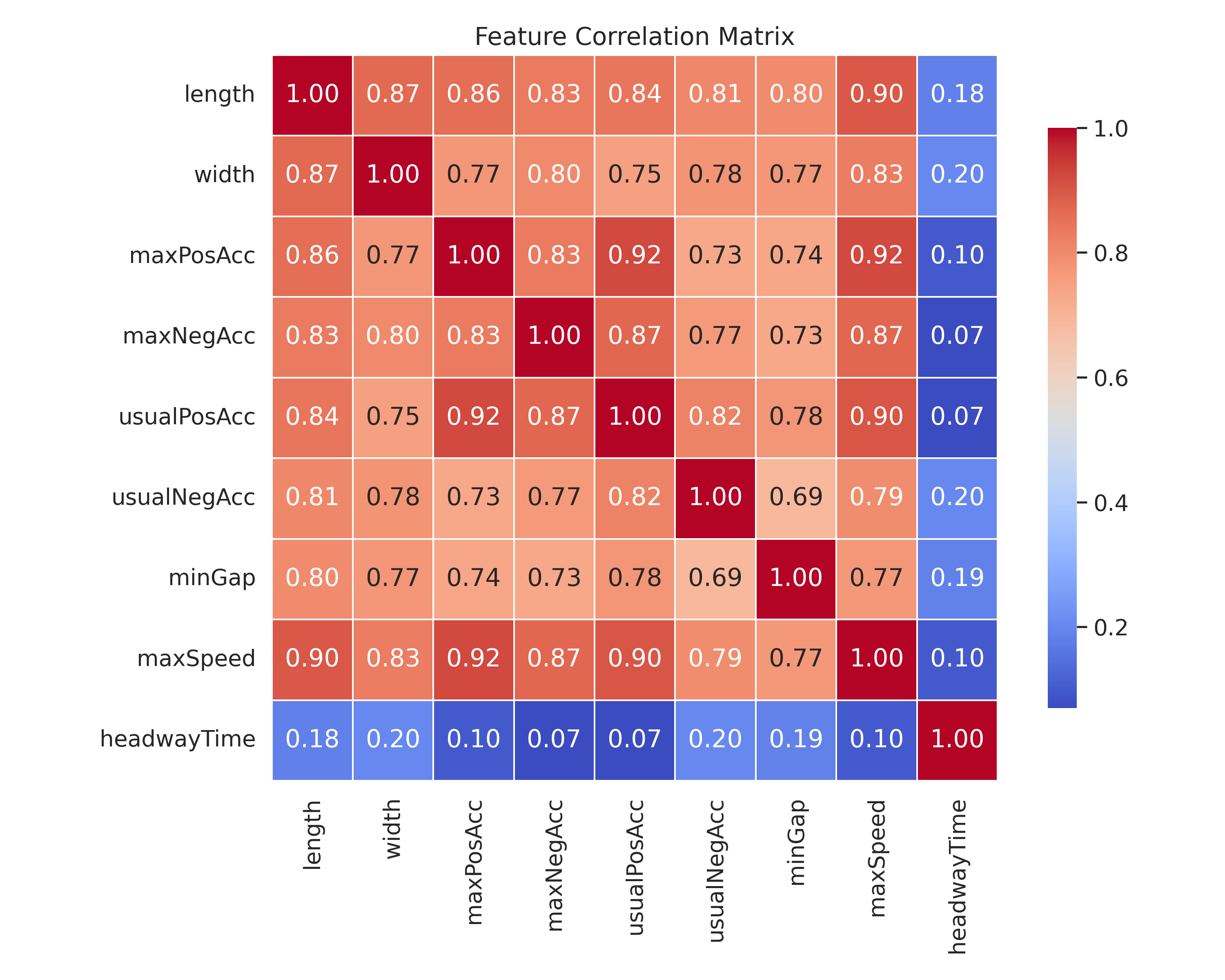}
    \caption{Correlation matrix for cMALC-D generated curriculum on the JN environment.}
    \label{fig:JN_correlation}
\end{figure}

\subsection{LLM Prompts}
\begin{longtable}{p{\textwidth}}
  \caption{The following table contains the prompts used for the LLM, both initially and after receiving data from the environments. }\label{fig:visual_hallucination}\\
  \fontsize{9.0pt}{\baselineskip}\selectfont
  \linespread{0.9}\selectfont
  \resizebox{\textwidth}{!}{
  \begin{mybody}
  \scriptsize{
  \textbf{This is the initial prompt given to the LLM.} \\\\
    \textbf{System:} 
    You are a curriculum designer for traffic simulations. Your goal is to generate a curriculum for training an agent to optimize traffic flow. This curriculum needs to test the agent's ability to handle various traffic conditions. Generate one plausible set of car parameters within the given bounds:\\

- length: (1.0-10.0)\\
- width: (1.0-5.0)\\
- maxPosAcc: (0.5-5.0)\\
- maxNegAcc: (0.5-5.0)\\
- usualPosAcc: (1.0-5.0)\\
- usualNegAcc: (1.0-5.0)\\
- minGap: (1.0-10.0)\\
- maxSpeed: (3.0-15.0)\\
- headwayTime: (1-5, integer)\\

Output a single JSON object of parameter values. \\ \\ 

\textbf{LLM:} ...
    }
  \end{mybody}
  }
  \resizebox{\textwidth}{!}{
  \begin{mybody}
  \scriptsize{
  \textbf{This is the prompt given to the LLM at all later timesteps, after having results from previous contexts} \\\\
  \textbf{System:} You are a curriculum designer for traffic light simulation. Your goal is to generate a curriculum for training an agent to optimize traffic flow. This curriculum needs to test the agent's ability to handle various traffic conditions.\\

Use the past trial data to propose the next car configuration for training.\\

    Analyze these past car parameter trials and determine how to generate the next task: \\

[results given in json format here] \\

Car Performance Assessment: \\
1. What parameter combinations were successful?\\
2. What weaknesses should be addressed?\\
3. What logical evolutions can we make?\\
4. Suggest specific values or parameter patterns to evolve the curriculum.\\

Format your response as:\\
- 1-2 sentences summarizing key insights\\
- Then "NEXT TASK SUGGESTION:" followed by a JSON object of new car parameters satisfying the bounds:\\

- length: (1.0-10.0)\\
- width: (1.0-5.0)\\
- maxPosAcc: (0.5-5.0)\\
- maxNegAcc: (0.5-5.0)\\
- usualPosAcc: (1.0-5.0)\\
- usualNegAcc: (1.0-5.0)\\
- minGap: (1.0-10.0)\\
- maxSpeed: (3.0-15.0)\\
- headwayTime: (1-5, integer)\\

Please generate your insights and new parameters. \\ \\

\textbf{LLM:} ...
    }
  \end{mybody}
  }
\end{longtable}

\end{document}